\def\year{2022}\relax
\documentclass[letterpaper]{article} 
\usepackage{aaai22}  
\usepackage{times}  
\usepackage{helvet}  
\usepackage{courier}  
\usepackage[hyphens]{url}  
\usepackage{graphicx} 
\usepackage{natbib}  
\usepackage{caption} 
\DeclareCaptionStyle{ruled}{labelfont=normalfont,labelsep=colon,strut=off} 
\usepackage{algorithm}
\usepackage{algorithmic}

\usepackage[utf8]{inputenc} 
\usepackage[T1]{fontenc}    
\usepackage{hyperref}       
\usepackage{url}            
\usepackage{booktabs}       
\usepackage{amsfonts}       
\usepackage{nicefrac}       
\usepackage{microtype}      
\usepackage{color}
\usepackage{xcolor}         

\usepackage{caption}
\usepackage{booktabs}
\usepackage{url}
\usepackage{todonotes}
\usepackage{multirow}
\usepackage{tikz}

\usepackage{amsmath}
\usepackage{amssymb}
\usepackage{amsthm}
\usepackage{amsfonts}
\usepackage{thmtools}		
\usepackage{mleftright}
\usepackage{stmaryrd}
\usepackage{nicefrac}

\usepackage{subcaption}
\usepackage{wrapfig}

\usepackage{microtype}
\usepackage{graphicx}
\usepackage{booktabs} 
\usepackage{amssymb}
\usepackage{amsmath}
\usepackage{float}
\usepackage{multirow}
\usepackage{natbib}
\usepackage{hyperref}
\usepackage{bbold}

\usepackage{xcolor,colortbl}
\definecolor{extreme}{gray}{0.85}
\definecolor{bad}{gray}{0.95}
\usepackage{amsmath}
\usepackage{caption}
\usepackage{multicol}
\usepackage{algorithm}
\usepackage{algorithmic}
\def\calV{{\mathcal{V}}}
\renewcommand{\algorithmiccomment}[1]{\bgroup\hfill//~#1\egroup}

\usepackage{newfloat}
\usepackage{listings}
\floatstyle{ruled}
\newfloat{listing}{tb}{lst}{}
\floatname{listing}{Listing}
\title{Graph Representation Learning with Individualization and Refinement}
\author {
    Mohammed Haroon Dupty, 
    Wee Sun Lee 
}
\affiliations{
	National University of Singapore \\
	\{dmharoon, leews\}@comp.nus.edu.sg
	
}
\usepackage{bibentry}

\begin{document}

\maketitle

\begin{abstract}
	Graph Neural Networks (GNNs) have emerged as prominent models for representation learning on graph structured data. GNNs follow an approach of message passing analogous to 1-dimensional Weisfeiler Lehman (1-WL) test for graph isomorphism and consequently are limited by the distinguishing power of 1-WL. More expressive higher-order GNNs which operate on k-tuples of nodes need increased computational resources in order to process higher-order tensors. Instead of the WL approach, in this work, we follow the classical approach of \emph{Individualization and Refinement} (IR), a technique followed by most practical isomorphism solvers. Individualization refers to artificially distinguishing a node in the graph and refinement is the propagation of this information to other nodes through message passing. We learn to adaptively select nodes to individualize and to aggregate the resulting graphs after refinement to help handle the complexity. Our technique lets us learn richer node embeddings while keeping the computational complexity manageable. Theoretically, we show that our procedure is more expressive than the 1-WL test. Experiments show that our method outperforms prominent 1-WL GNN models as well as competitive higher-order baselines on several benchmark synthetic and real datasets. Furthermore, our method opens new doors for exploring the paradigm of learning on graph structures with individualization and refinement.
\end{abstract}
\section{Introduction} 
Graphs are one of the most useful data structure in terms of representing real world data like molecules, social networks and citation networks. Consequently, representation learning~\citep{bengio2013representation,hamilton2017inductive,hamilton2017representation} on graph structured data has gained prominence in the machine learning community. 
In recent years, Graph Neural Networks (GNNs)~\citep{kipf2016semi,defferrard2016convolutional,gilmer2017neural,wu2020comprehensive} have become models of choice in learning representations for graph structured data. GNNs operate locally on each node by iteratively aggregating information from its neighbourhood in the form of messages and updating its embedding based on the messages received. This local procedure has been shown to capture some of the global structure of the graph. However, there are many simple topological properties like detecting triangles that GNNs fail to capture~\citep{chen2020can,knyazev2019understanding}. 

Theoretically, GNNs have been shown to be no more powerful than 1-dimensional Weisfeiler Lehman (1-WL) test for graph isomorphism~\citep{xu2018powerful,morris2019weisfeiler} in terms of their distinguishing capacity of non-isomorphic graphs. 1-WL is a classical technique of iteratively coloring vertices by injectively hashing the multisets of colors of their adjacent vertices. Although this procedure produces a stable coloring, which can be used to compare graphs, it fails to uniquely color many non-symmetric vertices. This is precisely the reason for the failure of 1-WL and thereby GNNs, to distinguish many simple graph structures~\citep{kiefer2020power,arvind2020weisfeiler}. Further works on improving the power of GNNs have tried to incorporate higher-order WL generalizations in the form of $k$-order GNNs~\citep{morris2019weisfeiler,maron2019provably,morris2020weisfeiler}, These $k$-order GNNs jettison the local nature of 1-WL and operate on $k$-tuples of vertices and hence are computationally demanding. However, this suggests techniques for learning better node representations can be found in classical algorithms used to detect graph isomorphisms.

Another possible approach for learning node features is the \emph{individualization and refinement} (IR) paradigm which provides a useful toolbox for testing graph isomorphism. In fact, all state-of-the-art isomorphism solvers follow this approach and are fairly fast in practice~\citep{mckay1981practical,darga2004exploiting,junttila2011conflict,mckay2014practical}. IR techniques aim to assign a distinct color to each vertex such that the coloring respects permutation invariance and is \emph{canonical}, a unique coloring of its isomorphism class (no two non-isomorphic graphs get the same coloring). 
This is achieved first by \emph{individualizion}, a process of recoloring a vertex in order to artificially distinguish it from rest of the vertices. Thereafter, a color \emph{refinement} procedure like the 1-WL message passing is applied to propagate this information across the graph. This process of IR is repeated until each vertex gets a unique color. But the coloring is not \emph{canonical} yet. To preserve permutation invariance, whenever a vertex is individualized, we have to individualize, and thereafter refine, all other vertices with the same color as well.
As each individualization of a vertex gives a different final discrete coloring, this generates a search-tree of colorings where each individualization forms a tree-node. The tree generated is \emph{canonical}. The coloring of the graph at each leaf can also be used to label the graph, and the graph with the largest label can be used as a canonical graph. However, the size of search-tree grows exponentially fast and isomorphism-solvers prune the search-tree heavily by detecting symmetries like automorphisms, and by other hand-crafted techniques. 

In this paper, we propose to improve representations learnt by GNNs by incorporating the inductive biases suggested by \emph{individualization and refinement} in updating the node embeddings. Unlike isomorphism-solvers, it is not desirable to check for automorphisms to prune the search-tree from a learning perspective. For computational efficiency, we restrict our search to individualizing $k$ nodes in each iteration. In order to restrict the search from exponentially blowing up, we take the approach similar to beam search and reduce the $k$ refined graphs to a single representative graph and repeat the individualization and refinement process. This simple technique lets us learn richer node embeddings while keeping the computational complexity manageable.
We validate our approach with experiments on synthetic and real datasets, where we show our model does well on problems where 1-WL GNN models clearly fail. Then we show that our model outperforms other prominent higher-order GNNs by a substantial margin on some real datasets.

\section{Related Work}
Over the last few years, there have been considerable number of works on understanding the representative power of GNNs in terms of their distinguishing capacity of non-isomorphic graphs~\citep{xu2018powerful,morris2019weisfeiler, chen2019equivalence,dehmamy2019understanding,srinivasan2019equivalence,loukas2019graph,barcelo2019logical}. These works have established that the GNNs are no more expressive than 1-WL kernels in graph classification~\citep{shervashidze2011weisfeiler}. Furthermore, ~\citet{chen2020can} have shown that many of the commonly occurring substructures cannot be detected by GNNs. One example is counting the number of triangles which can be easily computed from third power of the adjacency matrix of the graph~\citep{knyazev2019understanding}.

To improve the expressive power of GNNs, multiple works have tried to break away from this limitation of local message passing of 1-WL. Higher-order GNNs which can be seen as neural versions of $k$-dimensional WL tests have been proposed and studied in series of works~\citep{maron2018invariant, morris2019weisfeiler,maron2019provably, morris2020weisfeiler}. Although, these models are provably more powerful than 1-WL GNNs, they suffer from computational bottlenecks as they operate on $k$-tuples of nodes.~\citet{maron2019provably} proposed a computationally feasible model but was limited to 3-WL expressivity. Note that 1-WL and 2-WL have same expressivity ~\citep{maron2019provably}.
Other line of works on improving the expressivity of GNNs tend to introduce extra features which can be computed by preprocessing the graphs in the form of distance encoding~\citep{li2020distance} and subgraph isomorphism counting~\citep{bouritsas2020improving}. These techniques help in practice but fall short in the quest of better algorithms to extract such information by improved learning algorithms. 

The main problem of message passing in GNNs is that nodes fail to uniquely identify if the successive messages received are from the same or different nodes. In order to address this issue, ~\citet{sato2020random} and \citet{dasoulas2019coloring} have proposed to introduce features with random identifiers for each node. These models, though showing some benefit, can only maintain permutation invariance in expectation. Furthermore, structural message passing in the form of matrices with unique identifiers may help as shown in~\citet{vignac2020building} but its fast version has the same expressivity as in~\citet{maron2019provably}.

In this work, we aim to address some these concerns by leveraging the  \emph{individualization and refinement} paradigm of isomorphism solvers which are often fast in practice. We make a few approximations by individualizing fixed number of nodes and keeping a single representative graph in each iteration to manage the computational complexity. To avoid loss in accuracy, we use learning both for adaptively selecting the nodes to individualize and for merging the graphs from data. Our approach provides flexible trade off between robustness and speed of the model using the number of nodes to individualize as a hyperparameter which can be tuned using training data. 

\section{Preliminaries}\label{sec:prelims}
Let $G=(\mathcal{V}, \mathcal{E})$ be a graph with vertex set $\mathcal{V}$ and edge set $\mathcal{E}$. 
Two graphs $G, G'$ are \emph{isomorphic}, if there exists an adjacency-preserving bijective mapping $f: \calV_G  \rightarrow \calV_{G'}'$, i.e. $(v,u) \in \mathcal{E}$ iff $(f(v),f(u)) \in \mathcal{E}'$.  An \textit{automorphism} of $G$ is an isomorphism that maps $G$ onto itself. Intuitively, two vertices can be mapped to each other via an automorphism if they are structurally indistinguishable in the graph. 
A prominent way of identifying isomorphism is by coloring the nodes based on the structure of the graph in a permutation invariant manner and comparing if two graphs have the same coloring. Formally, a \emph{vertex coloring} $\pi$ is a surjective  
function $ \pi : \calV \rightarrow \mathrm{N}$ which assigns each vertex of the graph to a color (natural number).
A graph $G$ is called \emph{colored graph} $(G,\pi)$, if $\pi$ is a coloring of $G$. 
Given a graph $G$, a \emph{cell} of $\pi$ is the set of vertices with a given color. Vertex coloring partitions $\calV$ into \emph{cells} of color classes and hence is often called a \emph{partition}. If any two vertices of the same color are  adjacent to  the  same number  of vertices  of each  color, then such a coloring is called \emph{equitable coloring}, which cannot be further refined. If $\pi,\pi'$ are equitable colorings of a graph, then $\pi'$ is \emph{finer than or equal to} $\pi$, if 
$\pi(v)<\pi(w)\Rightarrow\pi'(v)<\pi'(w)$ for all $v,w\in V$. This implies that each cell of $\pi'$ is a subset of a cell of~$\pi$,
but the converse is not true. A ~\emph{discrete coloring} is a coloring where each color \emph{cell} has a single vertex i.e. each vertex is assigned a distinct color.

\subsection{Vertex refinement or 1-WL test}
1-dimensional Weisfeiler Lehman coloring is a graph coloring algorithm used to test isomorphism between graphs. Initially, all vertices are labeled uniformly with the same color and then are refined iteratively based on the local neighbourhood of each vertex. In each iteration, two vertices are assigned different colors if the multisets of their colored neighbourhoods are different. Specifically, if $\pi^t(v)$ is the color of vertex $v$ at time step $t$, then the colors are refined with the following update: $\pi^{t+1}(v) = \mathrm{HASH}\Big(\pi^{t}(v), \  \{\!\!\{ \pi^{t}(u), u\in N(v)\}\!\!\}\Big)$, where $\{\!\!\{ \}\!\!\}$ denotes a multiset and $N(v)$ is the neighbourhood of $v$. This procedure produces an equitable coloring where no further refinement is possible and the algorithm stops. This is a powerful technique of node coloring but has been shown to be restricted for many classes of graphs which cannot be distinguished by 1-WL refinement~\citep{kiefer2020power,arvind2020weisfeiler,chen2020can}. 

\subsection{Individualization and refinement}
Most of the present graph isomorphism solvers i.e. \texttt{Nauty}~\citep{mckay1981practical}, \texttt{Traces}~\citep{mckay2014practical}, etc. are based on a coloring paradigm called \emph{individualization and refinement}. In practice, these solvers are quite fast even though they can take exponential time in worst case. A comprehensive explanation of the of individualization and refinement algorithms can be found in~\citet{mckay2014practical}. Below we give a brief description.

\begin{figure}
	\centerline{
		\includegraphics[width=0.9\linewidth]
		{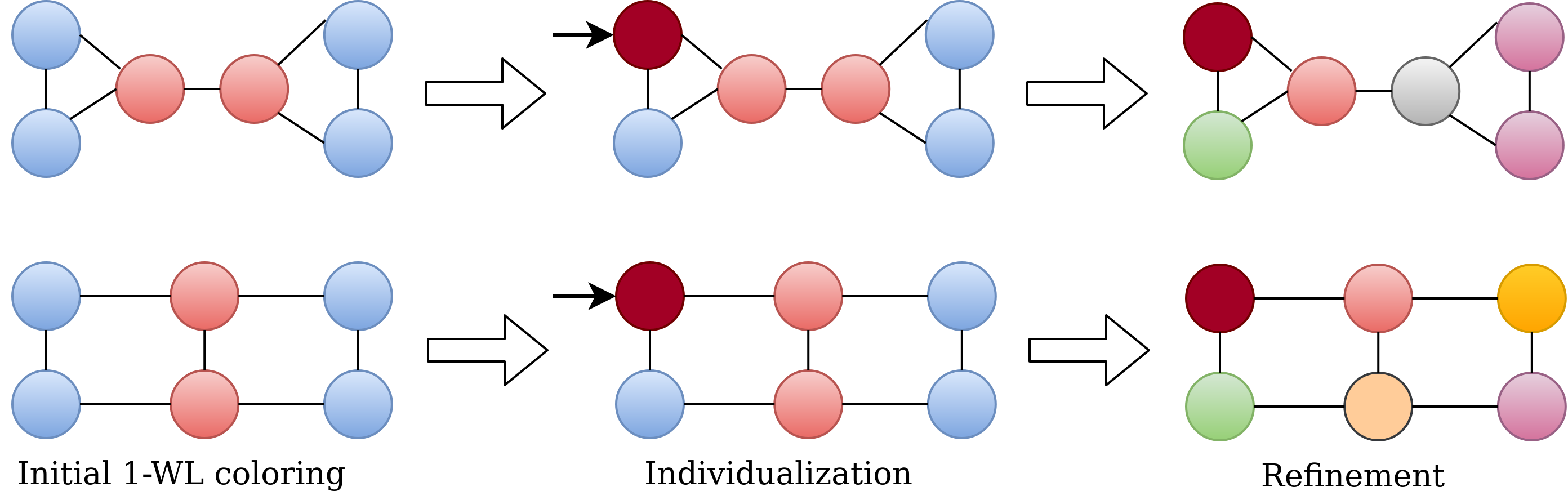}}
	\caption{\footnotesize Example graphs where 1-WL cannot distinguish two non-isomorphic graphs. Initially, 1-WL (or GNN) message passing refines the graphs into equitable partition of same two color \emph{cells} and hence the  graphs are indistinguishable. In the step of \emph{individualization}, one of the blue vertex is distinguished from other vertices by recoloring it (pointed by black arrows). Subsequently, 1-WL \emph{refinement} i.e. message passing, produces distinct multiset of colors for the two graphs, thereby distinguishing them as non-isomorphic.}
	\label{fig:individualization}	
	\vskip -0.1in

\end{figure}

Individualization refers to picking a vertex among vertices of a given color and distinguishing it with a new color. Once a vertex is distinguished from the rest, this information can be propagated to the other nodes by 1-WL message passing. An example is shown in the Figure~\ref{fig:individualization}, where initially 1-WL coloring is unable to distinguish the two graphs. Individualizing one of the blue colored vertices and further refinement produces  equitable coloring of the graphs such that they become distinguishable. 
\paragraph{}
The procedure to generate a \emph{canonical} coloring of the graph is as follows. Initially, vertex refinement is used to get the equitable coloring of the graph. This would partition the graph into a set of color cells. Then one of the color cells called \emph{target cell} is chosen and a vertex from it is reassigned a new color. This is propagated across the graph till a further refined equitable coloring of the graph is obtained. But this refinement comes at a cost. This can only be done in a permutation invariant manner if all the vertices of the chosen target cell are individualized and thereafter refined. If we chose a target cell of size $k$ i.e. the chosen color has been assigned to $k$ nodes in the graph, then after individualization and subsequent refinement, we would have $k$ colored graphs which are \emph{finer} than the one before individualization. Note that these $k$ graphs can still be not discrete and hence the process is repeated for each of $k$ refined graphs until all final graph colorings are discrete. This would take the shape of a search tree where tree nodes are graphs with equitable colorings. This search tree is \emph{canonical} to the isomorphism class of the graph 
i.e. the search tree is unique to the graph's isomorphism class. 
Finally, one of the leaves is chosen based on a predefined function which sorts all discrete colorings found as leaves of the search tree. 

Furthermore, vertices belonging to the same automorphism group in the graph always get the same color in the initial equitable coloring and all of them induce the same refined coloring of the graph when individualized. Intuitively, this is because there is no structural difference between the vertices of the same automorphism group. For example, in Figure~\ref{fig:individualization}, individualizing any of the blue vertices would produce the same set of node colorings. Therefore, isomorphism solvers prune the search tree by detecting automorphism groups. Effectively, even if the target cell is large, the branching can be reduced if automorphisms are detected.

\section{Proposed Method}
\label{sec:proposed_method}
\begin{figure*}[t]
	\centering
	\includegraphics[scale=0.065]{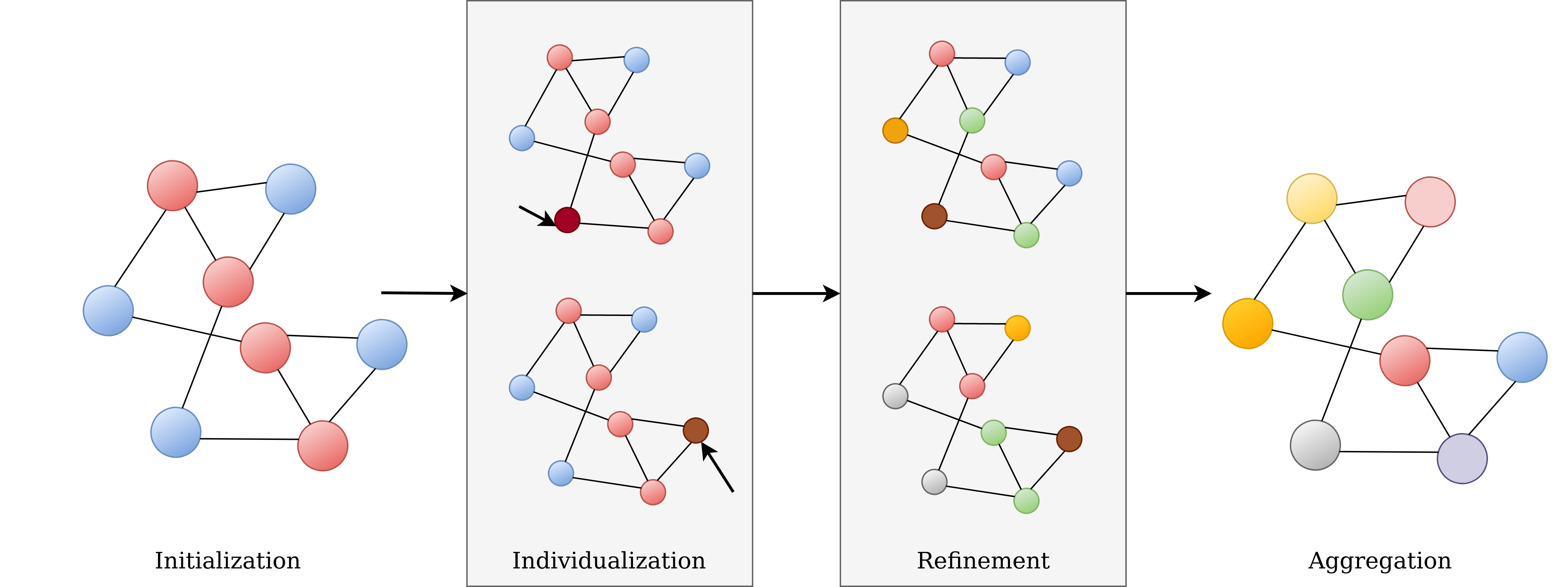}
	\caption{\footnotesize Given a graph, its node embeddings are initialized with a GNN. This partitions the graph into distinct color cells with nodes in a color cell having the same embedding. We select top-$k$ nodes nearest to \emph{target-cell} embedding 
		and separately \emph{individualize} them. 
		Subsequent \emph{refinement} by GNN message passing generates $k$ refined graphs. Graph-aware node embeddings across $k$ graphs are then aggregated into a single representative embedding. 
		This IR process is repeated $L$ times before a readout function is used to make the final prediction.}
	\label{fig:individualization2}
	\vskip -0.15in
\end{figure*}

GNNs learn node embeddings, by iteratively aggregating embeddings of its neighbours. In this work, we propose to leverage the \emph{individualization and refinement} paradigm in learning to distinguish nodes with similar embeddings. The key idea is to approximate the search process generated by IR based isomorphism solvers. To make it computationally feasible, we make neccessary approximations. A conceptual overview of our approach is shown in Figure~\ref{fig:individualization2} and pseudocode is shown in Algorithm~\ref{alg:GNN-IR}.\\
\textbf{Notations:} We refer to hidden vectors with $h_v$ for embeddings of node $v$ and $h_G$ for the aggregated embeddings of graph $G$, generated with appropriate global pooling function applied on node embeddings. We use $\mathbf{H} \in \mathbb{R}^{n\times d}$ to denote matrix of node embeddings. $h_v^l$  is the embedding at layer $l \in \{1\dots L\}$. Note that layer $l$ refers to one full iteration of individualization-refinement step. $h_v^0$ are initialized with node features or with a constant value in the absence of node features.  

\subsection{Initialization}
Graph Neural Networks (GNNs) 
are neural versions of 1-WL message passing and hence node embeddings would converge after iterating for fixed number of steps. At this stage, some of the nodes end up with same embeddings. This is equivalent to the equitable partitioning of the graph. This serves as the stable initialization of the node embeddings and partitions the graph akin to \emph{color cells}.

\subsection{Target cell selection}
After initial GNN refinement, the graph would be a multiset of color (embedding) classes. 
To break the symmetry, we need to choose a color cell called the \emph{target cell}, which contains the chosen set of nodes with the same color, in order to individualize each one of them. 
We approximate the \emph{target-cell} selection by choosing one of the node embeddings in the graph and selecting $k$ most similar nodes with the chosen embedding. The choice of the \emph{target cell} is important as it determines the breadth and the depth of the search tree and thereby the richness of the embeddings learnt. Usually, the isomorphism solvers use hand-engineered techniques to choose the target cell. Note that the size of the target cell itself may not matter much with respect to the refinement it induces on the graph. Some solvers like \texttt{Nauty} choose the smallest non-singleton cell whereas \texttt{Traces} choose the largest cell in order to shorten the depth of the search tree. 
This is one place where we can leverage learning from data. Given a dataset, the \emph{target cell} chosen should best fit the data. Let $\phi$ be a scoring function on nodes which can be used to select top-k indices for individualization in each iteration. If $\mathcal{I}^l$ is the set of nodes of the \emph{target cell} to be individualized at layer $l$ then,
\vskip -0.2in
\begin{equation}\label{eq:topk}
	\mathcal{I}^l = \texttt{Top-k-rank}(\{ \phi^l(h_v^l,G;\theta) , \forall v \in \mathcal{V}\})
\end{equation}
\vskip -0.05in
One simple method of implementing such a scoring function would be Top-K pooling as done in ~\citep{gao2019graph,lee2019self}. But this may not capture the best ~\emph{target cell} as these functions are only parameterized by a projection vector and do not provide specific inductive bias towards an end goal.
Intuitively, to bias the scoring function $\phi$ towards better \emph{target cells}, it should determine the \emph{cell} based on 
the current state of the graph i.e., the number and size of distinct embeddings (\emph{cells}) 
and the previous embeddings that have already been individualized i.e. history. 
Therefore, we model the target cell selector as a Recurrent neural network with a GRU cell as it can track previous states via its hidden state.

Precisely, in each step, the GRU cell outputs a projection vector which is used to rank the nodes based on the similarity of the nodes with the projection vector.  At iteration $l$, the GRU cell takes in two inputs, one the pooled feature of the graph $h_G^l$ and the other is the hidden state of the GRU from the previous time step. It then produces two embeddings, the updated hidden state and an output embedding. The output is considered as a surrogate for the target cell embedding on which all the nodes are projected and ranked based on the projection score. Let $\mathbf{p}_{tc}$ be the projection vector output by the GRU, $\mathrm{AGG}$ be an aggregator set function, $\mathbf{q}_{tc}$ be the hidden state vector and $\mathbf{H}$ be the set of node embeddings.
\begin{figure}
	\vskip -0.1in
	\resizebox{\linewidth}{!}{
		\begin{minipage}{0.55\textwidth}
			\LARGE
			\begin{algorithm}[H]
				\caption{The framework of GNN-IR}
				\label{alg:GNN-IR}
				\begin{algorithmic}[1]
					\STATE {\bfseries Input:} Graph $G =  (\mathcal{V},\mathcal{E})$; 
					\STATE $d$-layer GNN with 1-WL expressive power; 
					\STATE Target cell selector function $\phi$; 
					\STATE Individualization function $\psi$;
					\STATE Multiset aggregators AGG;
					MLP $f$;
					\STATE {\bfseries Output:} Refined embedding  for Graph $G$
					
					\STATE Initialize $\mathbf{H}^0 = \{h_v^0,  \forall v \in \mathcal{V}\}$ with constant value
					\STATE $\mathbf{H}^1$ $\leftarrow$ GNN($\mathbf{H}^0$)
					
					\FOR{$l=1$ {\bfseries to} $L$}
					\STATE 	$\mathcal{I}^l = \texttt{Top-k-rank}(\{ \phi^l(h_v^l,\mathbf{H}^l;\theta) , \forall v \in \mathcal{V}\})$
					
					\FOR{$i \in \mathcal{I}^l$}
					\STATE $\mathbf{H}^{l^i} \leftarrow \mathbf{H}^l$
					\STATE $h_i^{l^i} \leftarrow h_i^{l^i} \odot \psi(h_i^{l^i})$ \COMMENT{Individualization}   
					\STATE  $\tilde{\mathbf{H}}^{l^i}$ $\leftarrow$ GNN($\mathbf{H}^{l^i}$) \COMMENT{Refinement} 
					\STATE $\tilde{h}_G^{l^i}\leftarrow \text{AGG}(\tilde{\mathbf{H}}^{l^i})$ 
					\STATE $\tilde{\mathbf{H}}^{l^i} \leftarrow \{f(\tilde{h}_v^{l^i},\tilde{h}_G^{l^i};\theta), \forall v \in \mathcal{V}\}$
					\ENDFOR
					
					\STATE $\mathbf{H}^{l+1} \leftarrow \{\text{AGG}(\{\tilde{h}_v^{l^i} \; |\; \forall i \in \mathcal{I}\}) \; |\; \forall v \in \mathcal{V}\}$ \COMMENT{Merge}
					\ENDFOR
					
					\STATE {\bfseries Return} $\{\text{AGG}\big(\{h_v^l \; |\; l=0,\dots L \}\big) \;|\; \forall v \in \mathcal{V}\}$
				\end{algorithmic}
			\end{algorithm}
		\end{minipage}
	}
\vskip -0.1in
\end{figure}

Then the following rule is applied to rank the nodes and get the $k$ individualization indices.
\begin{align}
	h_G^l &= \mathrm{AGG}(\mathbf{H}^l)\label{eq:target_cell_1}\\
	\mathbf{p}_{tc}^l, \mathbf{q}_{tc}^l &= \mathrm{GRU}(h_G^l, \mathbf{q}_{tc}^{l-1})\label{eq:target_cell_2}\\
	\mathrm{score}^l &= \mathrm{tanh}(\frac{\mathbf{H}^l\mathbf{p}_{tc}^l}{\lVert \mathbf{p}_{tc}^l\rVert})\label{eq:target_cell_3}\\
	\mathcal{I} &= \text{top-k-rank}(\mathrm{score}, k) \label{eq:target_cell_4}\\
	\mathrm{Readout} & (\mathbf{p}_{tc}^l \odot \mathrm{score}^l)_{\mathcal{I}}
	\label{eq:target_cell_5}
\end{align}
where $\odot$ is elementwise product.
To select the nodes, the node embeddings are projected onto $\mathbf{p}_{tc}$ and a score is computed for each node. The top-k ranked indices are selected into the set $\mathcal{I}$. To be able to backpropagate through the scoring function, the top-k scores are multiplied with the projection vector and a Readout function is used to generate a separate feature vector which is concatenated with the final graph embedding before prediction.

\subsection{Individualization}
Now that we have the selected $k$ indices in $\mathcal{I}$, we individualize each of the $k$ nodes separately and branch out. Our individualization function $\psi$ is an MLP and acts only on the selected node embeddings. The node embeddings of $h_i, \;\forall i \in \mathcal{I}$ are updated by passing through individualization function;
thereafter, the embedding matrix of the graph is updated by the procedure of masking.
\begin{equation}
	\mathbf{Z}_{mask} = \mathbf{1} ; \; \; 
	\mathbf{Z}_{mask}[ i ] = \psi(h_i^{l_i}) ; \; \; 
	\mathbf{H} = \mathbf{H} \odot \mathbf{Z}_{mask} ; \; \; 
\end{equation}
This is repeated for each $i \in \mathcal{I}$ separately as we can only color one node at a time in the graph.
The node embedding matrix $\mathbf{H}$ is updated with the individualized node embedding. The next step is to refine the graph to propagate this information. 

\subsection{Refinement and Aggregation}
After each individualization, we run GNN again for a fixed number of steps till the node embeddings converge again. This process of \emph{Refinement} gives us $k$ refined graphs. Ideally, we should expand all the $k$ graphs further but this incurs extra computational cost which grows exponentially with depth. 

In cases where we need to search over a tree, one of the popular greedy heuristic is beam search, where the search tree of a fixed width is kept. Selection of best graph out of $k$ would require a value-function approximator which would score over each of the $k$ graphs. Such a function would be difficult to train given the limited amount of labeled data. As an approximation, we instead construct a multiset function to aggregate the $k$ embeddings of each node into a single embedding. In principle a universal multiset function approximator should be able to approximate selection as well. 
Also, here we need a set function of a set of sets i.e. set of $k$ graphs. For this we first construct graph-aware node embeddings by combining the node embeddings with the pooled embeddings of the same graph. 
Finally, node embeddings are aggregated across $k$ graphs to generate a new representative graph. 
Precisely, let $\tilde{\mathbf{H}}^{l^i}$ be the set of node embeddings after individualization and refinement of index $i$ in layer $l$. We first compute $\tilde{h}_G^{l^i}$, a representation for $G^{l^i}$ as 
\vskip -0.1in
\begin{equation}
	\tilde{h}_G^{l^i} =\sum_{v \in \mathcal{V}} \text{MLP} (\tilde{\mathbf{H}}^{l^i})
	\label{eq:aggregation_1}
\end{equation}\vskip -0.1in
Thereafter, we feed $\tilde{h}_G^{l^i}$ to an MLP  and add to each $\tilde{h}_v^{l^i}$ to get a graph-representative node embedding for the nodes of the $k$ refined graphs.
\vskip -0.1in
\begin{equation}
	\tilde{h}_v^{l^i} \leftarrow \tilde{h}_v^{l^i} + \text{MLP}( \tilde{h}_G^{l^i} )
	\label{eq:aggregation_2}
\end{equation}\vskip -0.05in
We then max pool the node embeddings across the $k$ graphs i.e. for a node $v$, we pool together $k$ embeddings $\tilde{h}_v^{l^i}$  to generate the aggregated new representation for node $v$.
\vskip -0.1in
\begin{equation}
	h_v^{l+1} = \max_{i \in \{1\dots k\}} \tilde{h}_v^{l^i}
	\label{aggregation_3}
\end{equation}
\vskip -0.1in
With the new representative graph embeddings, we repeat the IR procedure $L$ times before readout. 

\subsection{On aggregation of graphs and fixing k}
IR algorithms usually expand the tree until the colors are discrete. The colors are then used to label the leaves and the leaf with the largest label is selected. One way to simulate this is to learn a method of selecting the correct branch at each internal tree node in order to reach the correct leaf. 
Instead of selecting a branch, we learn to aggregrate the children of the internal nodes into a single node and find that aggregration works well in our experiments. 
In an ideal case where the multiset functions used for aggregration are universal approximators~\citep{zaheer2017deep,qi2017pointnet}, 
it would be able to learn to an approximate algorithm that selects one of the $k$ graphs, hence can learn to work as well as an algorithm that learns to select. We use a simpler aggregrator that is not necessarily a universal approximator. However, the use of max-pooling operator will likely make simulation of selection easier -- if all embeddings are positive, multiplying non-desired embeddings by a small number would result in max-pooling selecting the desired embedding.  



As for determining the value of $k$, different graphs can have different cell sizes and hence it is best to choose the value of $k$ by cross-validation. The best $k$ that fits the dataset is more likely to be the target cell size on average for a given layer. Cross-validation also helps when the cells consist of nodes from same automorphism group; for such cells, smaller value of $k$ would be sufficient. But as shown in Section~\ref{sec:analysis}, a larger value of $k$ would be more robust by including all nodes of the selected cell. 

\section{Analysis}\label{sec:analysis}
In this section, we discuss some of the properties of the proposed GNN model with \emph{individualization and refinement} as defined in Algorithm~\ref{alg:GNN-IR}, which we call as GNN-IR model. 

\textbf{Permutation invariance:} Each step of GNN-IR with arbitrary width $k$ preserves the property of permutation-invariance for a large class of graphs, though it may fail to preserve it for some graphs. 
All operators used in GNN-IR are either permutation-invariant operators on sets or those which operate on individual nodes. If the input graph has unique attributes for all nodes, then the node-operators will operate on same nodes irrespective of input node-permutations. Hence, GNN-IR is permutation-invariant for all graphs with unique node attributes. However, if the nodes are not distinguishable, then node-operators may operate on different nodes when the permutation of input nodes changes. The stage which decides which nodes are operated on is the  \emph{target cell} selection stage where top-$k$ nodes are selected for individualization.

Consider the graphs with no node-attributes. Let the initial coloring of the graph be $\pi=\{p_1,\dots,p_{\lvert \pi \rvert}\}$, where $p_i$ is the set of vertices with same color. For arbitrary width $k$, it is possible that not all vertices of a color-cell $p_i$ are included in the individualization set $\mathcal{I}$. Then, permutation-invariance is preserved if, upon individualization, all vertices of $p_i$ induce same refinement on the graph. Effectively, which $v\in p_i$ is included in $\mathcal{I}$ is irrelevant. This happens when $p_i$ forms an orbit i.e. all vertices of $p_i$ can be mapped to each other via an automorphism. Therefore, for arbitrary $k$, if all $p_i \in \pi$ are orbit cells \emph{i.e.} $\pi$ is an orbit-partition, then each step of GNN-IR preserves permutation invariance of the graph embeddings. 
For example, consider one of the graphs in Figure~\ref{fig:individualization}, where blue vertices form an orbit \emph{i.e.} all blue vertices are structurally equivalent, and hence individualization of any blue vertex of the graph will generate same refinement. Also, we show in Lemma 1 that, output of each step of GNN-IR remains an orbit-partitioned coloring and hence, permutation-invariance is preserved over multiple steps of GNN-IR.

\noindent\textbf{Lemma 1. } \emph{If the input to a GNN-IR step, which includes target-cell selection, individualization-refinement and aggregation, is an orbit-partitioned coloring, then the output will also be an orbit-partitioned coloring.}\\
\textit{Proof:} Proof is included in the Appendix.

\textbf{Expressive power:} We characterise the expressive power of GNN-IR in terms of its distinguishing capacity of the non-isomorphic graphs. 



\noindent\textbf{Proposition 1. } \emph{Assume we use universal set approximators in GNN-IR for target-cell selection, refinement and aggregation steps. GNN-IR is more expressive than all 1-WL equivalent GNNs i.e., GNN-IR can distinguish all graphs distinguishable by GNN and there exist graphs non-distinguishable by GNN which can be distinguished by GNN-IR.}\\
\textit{Proof sketch. }
A detailed proof is given in Appendix. Here, we give a brief sketch of the proof. First, we show that graphs distinguishable by GNNs generate orbit-partitions on 1-WL refinement. Hence, by Lemma 1, GNN-IR preserves permutation-invariance on these graphs and since, the first layer of GNN-IR is a GNN, they are distinguishable by GNN-IR. Next, we illustrate two graphs which are not-distinguishable by GNN and show how a step of GNN-IR including individualization-refinement and aggregation operators, distinguishes these two graphs. \hfill \qedsymbol 

\textbf{Runtime analysis:}
For bounded degree graphs, runtime of GNN for fixed iterations is $O(n)$. GNN-IR builds on GNN and in each IR step, GNN is run $k$ times. Assuming constant time aggregation with 1-layer MLPs, running GNN-IR for $L$ steps takes $O(Lkn)$.

\section{Experiments}\label{sec:experiments}
In this section, we report evaluation results of GNN-IR with multiple experiments. 

\noindent\textbf{Model architecture:} We follow the MPNN~\citep{gilmer2017neural} architecture in all our experiments. We first convolve with a 1-WL convolution operator, then update with a GRU cell~\citep{chung2014empirical} and for readout, we use either sum-pooling or set2set~\citep{vinyals2015order} function. For convolution, we use either GIN~\citep{xu2018powerful}, NNConv~\citep{gilmer2017neural} or PNA~\citep{corso2020principal} convolution operators depending on the dataset and availability of edge features. 
Also, we run GNN in each IR layer for $3$ steps and share parameters of the GNN in each iteration of IR which allows us to go deeper without increasing the parameters. We use 1-hidden layer MLPs in all set aggregators along with sum pooling. Also, in datasets with edge features, we consider edges as variables and readout from node and edge variables before final prediction. 
Code was implemented in Pytorch-Geometric~\citep{Fey/Lenssen/2019}.

\subsection{Counting Triangles}
\setlength{\columnsep}{8pt}
\definecolor{extreme}{gray}{0.85}
\definecolor{bad}{gray}{0.95}
\begin{wraptable}[]{R}{0.69\linewidth}
	\Large
	\vskip -0.1in
	\setlength{\belowcaptionskip} {-2pt}
	\setlength{\abovecaptionskip} {6pt}
	\centering		
	\resizebox{\linewidth}{!}{ 	\renewcommand{\arraystretch}{1.05}	
		\begin{tabular}{cccccc}
			\toprule
			& Train & \multicolumn{2}{c}{Test} & \multicolumn{2}{c}{Time}\\
			\cmidrule(lr){3-4}					\cmidrule(lr){5-6}
			&  & orig & large  & sec/ep & ratio\\
			\midrule
			\noalign{\vskip 1mm}
			GIN, top-k & $90${\tiny $\pm3$} & $47${\tiny$\pm2$} & $18${\tiny$\pm 1$} & - & - \\
			ChebyGIN & - & $64${\tiny$\pm5$} & $25${\tiny$\pm2$} & - & - \\
			GAT & $92${\tiny$\pm2$} & $50${\tiny$\pm1$} & $25${\tiny$\pm 1$}& - & - \\
			GIN* & $90${\tiny$\pm3$} & $47${\tiny$\pm2$} & $25${\tiny$\pm 1$}& 17.1 & 1 \\
			\midrule
			\noalign{\vskip 1mm}
			GNN - IR & & & & &\\
			\cmidrule{1-1}
			$L=1$\;\; $k=2$ & {90}{\tiny$\pm3$} & {76}{\tiny$\pm3$}&   {28}{\tiny$\pm2$}& 42.1 & 2.46\\
			$L=1$\;\; $k=4$ & {92}{\tiny$\pm3$} & {86}{\tiny$\pm1$}&   {34}{\tiny$\pm2$}& 51.7 & 3.02 \\					
			$L=2$\;\; $k=2$ & {86}{\tiny$\pm2$} & {78}{\tiny$\pm2$}&   {28}{\tiny$\pm4$}& 57.0 & 3.33 \\				
			$L=2$\;\; $k=4$ & \cellcolor{bad}{98}{\tiny$\pm1$} & \cellcolor{bad}{97}{\tiny$\pm1$}&  \cellcolor{bad}{41}{\tiny$\pm2$}& 80.8 & 4.73
			\\
			$L=3$\;\; $k=2$ & \cellcolor{bad}{93}{\tiny$\pm3$} & \cellcolor{bad}{91}{\tiny$\pm2$}&  \cellcolor{bad}{46}{\tiny$\pm2$}& 85.6 & 5.00
			\\
			$L=3$\;\; $k=4$ & \cellcolor{extreme} {99}{\tiny$\pm0$} & \cellcolor{extreme}{99}{\tiny$\pm1$}&  \cellcolor{extreme} \cellcolor{extreme}{51}{\tiny$\pm1$}& 98.7 & 5.77 \\
			\bottomrule
	\end{tabular}}
	\caption{\footnotesize Accuracy on counting the number of triangles in TRIANGLES dataset. $L$ is the number of layers of IR and $k$ is the width parameter. $^*$Our baseline model with $L=0, k=0$.}
\label{tab:Trianles}
	\vskip -0.1in
\end{wraptable}	
\paragraph{}
Counting triangles is a simple task and has an analytic solution of trace$(A^3)/6$, where $A$ is the adjacency matrix of the graph. But, 1-WL GNN models provably cannot count triangles~\citep{chen2020can}. Therefore, we evaluate GNN-IR on a publicly available synthetic dataset TRIANGLES~\citep{knyazev2019understanding} of $45000$ graphs where the task is to count the number of triangles. The dataset comes with 4 splits, training ($30000$), validation ($5000$), test-original ($5000$) and test-large ($5000$). The first three sets all have graphs with nodes $< 25$ and the test-large set has as many as $100$ nodes. The test-large is challenging and it tests the generalization ability of the model. 
We compare with baseline 1-WL GNN models, GAT, GIN and ChebyGIN~\citep{knyazev2019understanding} which is a more powerful GNN.

Table~\ref{tab:Trianles} shows the accuracy of the models on both the test sets. To analyse the effect of depth and width of IR, we report results for various values of $L$, number of IR layers and $k$, number of individualizations in each layer. As shown in results, just one layer of IR with $k=2$ individualizations significantly increases the accuracy from $47$ to $76$ for Test-orig. Furthermore, a clear pattern emerging from the results is that accuracy and generalization improves for longer depth $L$ and larger width $k$, with highest score saturating at $L=3$ and $k=4$. 
We also report the extra time taken by each IR layer, which includes $3$ steps of GNN per individualization, and compare it with GIN as it is our base GNN. The time ratio w.r.t GIN shows that with more IR layers and increased width $k$, the amount of computation also increases; but the increase is only linear both in $k$ and $L$.


\subsection{Recognizing Circulant  Skip  Links  (CSL) graphs }
\begin{wraptable}{R}{0.73\linewidth}
    \LARGE
	\vskip -0.15in
	\setlength{\belowcaptionskip} {-2pt}
	\setlength{\abovecaptionskip} {6pt}
	\resizebox{\linewidth}{!}{\renewcommand{\arraystretch}{1}
		\begin{tabular}{cccccc}
			\toprule
			& mean & median & max  & min & std\\
			\midrule
			\noalign{\vskip 1mm}
			GIN$^*$ & $10$ & $10$ & $10$& $10$ & $0$ \\
			RP-GIN & $37.6$ & $43.3$ & $53.3$ & $10$ & $12.9$ \\
			3WLGNN & $97.8$ & $-$ & $100.0$ & $30$ & $10.9$ \\
			\midrule
			\noalign{\vskip 1mm}
			GNN - IR & & & & &\\
			\cmidrule{1-1}
			$L-5$\;\; $k-04$ & $50$ & $60$ & $70$ & 
			$10$ & $23.45$
			\\
			$L=3$\;\; $k=08$ & $74$ & $70$ & $90$ & $60$ & $15.17$ \\				
			$L=5$\;\; $k=08$ & \cellcolor{bad}{$82$} & \cellcolor{bad}{$80$} & \cellcolor{bad}{$100$} & 
			\cellcolor{bad}{$70$} & $13.04$
			\\
			$L=2$\;\; $k=16$ & \cellcolor{bad}{$82$} & \cellcolor{bad}{$80$} & \cellcolor{bad}{$90$} & \cellcolor{bad}{$80$} & $4.47$ 
			\\		
			$L=3$\;\; $k=16$ & \cellcolor{extreme}{$96.67$} & \cellcolor{extreme}{$100.0$} &  \cellcolor{extreme}{$100.0$}& \cellcolor{extreme}{$90.0$} & $4.7$ \\
			$L=4$\;\; $k=16$ & \cellcolor{extreme}{$98.67$} & \cellcolor{extreme}{$100.0$} &  \cellcolor{extreme}{$100.0$}& \cellcolor{extreme}{$96.67$} & $1.8$ \\
			\bottomrule
	\end{tabular}}
	\caption{\footnotesize Graph classification on the CSL dataset. $^*$Our baseline model with $L=0, k=0$.}
	\label{tab:CSL}
	\vskip -0.1in
\end{wraptable}
\paragraph{}

\definecolor{extreme}{gray}{0.85}
\definecolor{bad}{gray}{0.95}
\newcommand\crule[3][black]{\textcolor{#1}{\rule{#2}{#3}}}
\begin{table*}[t]
    \LARGE
	\begin{minipage}{0.37\textwidth}
		\centering
		\resizebox{\columnwidth}{!}{
			\begin{tabular}{lcc}
				\toprule
				Data set & ZINC 10K & ALCHEMY 10K  \\
				& mae & $\;\;\;\;$ mae $\;\;\;\;\;\;\;\;$ log-mae\\
				\midrule
				GINE-$\epsilon$    & 0.278{\scriptsize $\pm 0.022$} & 0.185{\scriptsize $\pm 0.007$} -1.864{\scriptsize $\pm 0.062$}\\
				2-WL-GNN & 0.399{\scriptsize $\pm 0.006$} & 0.149{\scriptsize $\pm 0.004$} -2.609{\scriptsize $\pm 0.029$}\\
				$\delta$-2-GNN & 0.374{\scriptsize $\pm 0.022$} & {\bfseries 0.118{\scriptsize $\pm 0.001$}} -2.679{\scriptsize $\pm 0.044$} \\
				$\delta$-2-LGNN    & 0.306{\scriptsize $\pm 0.044$} & 0.122{\scriptsize $\pm 0.003$} -2.573{\scriptsize $\pm 0.078$}       \\
				DGN & 0.168{\scriptsize $\pm 0.010$} & 
				- \\
				PNA & 0.187{\scriptsize $\pm 0.010$} & 
				0.162{\scriptsize $\pm 0.005$} 
				-2.033{\scriptsize $\pm 0.054$} \\
				\midrule
				GNN-IR & {\bfseries 0.137}{\scriptsize $\pm 0.010$} & 
				0.119{\scriptsize $\pm 0.002$} 
				\bfseries -2.742{\scriptsize $\pm 0.070$}\\				
				\bottomrule
			\end{tabular}
		}
		\caption{\footnotesize MAE on ZINC10K and ALCHEMY10K \label{table3_zincalchemy}}
	\end{minipage}
	\hfill
	\begin{minipage}{0.19\textwidth}
		\centering
		\resizebox{\columnwidth}{!}{
			\begin{tabular}{lc}
				\toprule
				Data set & QM9 (mae)  \\
				\midrule
				GINE-$\epsilon$  &  0.081 {\scriptsize $\pm 0.003$} \\	
				MPNN & 0.034 {\scriptsize $\pm 0.001$}   \\	
				$1$-$2$-GNN &  0.068   {\scriptsize $\pm 0.001  $}  \\   
				$1$-$3$-GNN &  0.088 {\scriptsize $\pm 0.007$}  \\
				\textsf{$1$-$2$-$3$-GNN} & 0.062  {\scriptsize $\pm 0.001$} \\	
				$3$-IGN &  0.046 {\scriptsize $\pm 0.001$}  \\	
				$\delta$-$2$-LGNN & 0.029 {\scriptsize $\pm 0.001$} \\
				LRBP-net &  0.027 {\scriptsize $\pm 0.001$}  \\
				\midrule
				GNN-IR & \bfseries 0.020 {\scriptsize $\pm 0.001$}	\\
				\bottomrule
			\end{tabular}
		}
		\caption{\footnotesize Graph regression on QM9 dataset in std mae.\label{table3_qm9}}
	\end{minipage}
	\hfill
	\begin{minipage}{0.4\textwidth}
		\centering
		\resizebox{\columnwidth}{!}{
			\begin{tabular}{cccccc}
				\toprule
				Data set &  $O(.)$  & \multicolumn{2}{c}{ZINC10K} & \multicolumn{2}{c}{ALCHEMY10K}\\
				\cmidrule(lr){3-4}					\cmidrule(lr){5-6}
				&  & Time-ratio & mae  & Time-ratio & mae\\
				\midrule
				\noalign{\vskip 1mm}
				GINE (1-WL) &  $O(n)$  &  1.0  &  0.27  &  1.0  &  0.18  \\
				2-WL-GNN &  $O(n^2)$  &  17.11  &  0.39  &  7.33  &  0.14  \\
				$\delta$ -2-WL-GNN &  $O(n^2)$  &  21.86  &  0.37  &  10.71  &  0.11  \\
				\midrule
				GNN - IR &  $O(Lkn)$  & & & &\\
				\cmidrule{1-1}
				L=1 \;\;  k=2  &  $O(2n)$  &  4.61 &  0.16  &  2.38  &  0.13  \\
				L=1 \;\;  k=4  &  $O(4n)$  &  6.87  &  0.14  &   4.47  &  0.12  \\
				L=2 \;\;  k=2  &  $O(4n)$  &  7.36   &  0.13  &  4.72  &  0.12  \\
				L=2 \;\;  k=4  &  $O(8n)$  &   11.56  &  0.13  &  7.22  &  0.11  \\
				\bottomrule
			\end{tabular}
		}
		\caption{\footnotesize Runtime analysis: Time-ratio compared to GINE (1-WL).  \label{table3_runtime}}
	\end{minipage}
\end{table*}
	
A type of graphs that 1-WL GNN models cannot classify are regular graphs that do not provide any information in node degrees. In order to evaluate GNN-IR for highly regular graphs, we use Circulant Skip Links (CSL) dataset released by~\citep{murphy2019relational}. A CSL graph $\mathcal{G}_{skip} (M, R)$ is a 4-regular graph with $\{0,1,\dots M-1\}$ vertices with an edge between vertices $R$ distance from each other. 
The dataset consists of $150$ graphs with $10$ classes ($R$). 
We evaluate GNN-IR with CSL dataset with $5$-fold cross-validation as in~\citep{murphy2019relational} and report results along with 1-WL and more expressive models like RP-GNN~\citep{murphy2019relational} and 3-WL GNN. 

Since CSL graphs have highly regular structure, the number of individualizations needed to break the symmetry is high. Table~\ref{tab:CSL} shows that 1-WL GIN model fails whereas there is clear improvement by GNN-IR. Note that higher values of $k$ result in much better gain compared to increase in number of IR layers $L$. This suggests that with larger width $k$, either target color-cells with larger width are more helpful in breaking the symmetry or more number of smaller color-cells included for individualization helps in including the optimal target-cell in the selected $k$ nodes. 

\subsection{Real world benchmarks}
We now evaluate GNN-IR on real-world datasets on graph regression task. 
We use ZINC10K~\citep{jin2018junction,dwivedi2020benchmarking}, ALCHEMY10K~\citep{chen2019alchemy} and QM9~\citep{ruddigkeit2012enumeration,ramakrishnan2014quantum} datasets, since these datasets are used by recent prominent models~\citep{morris2020weisfeiler,corso2020principal} and to compare with the neural versions of higher-order GNNs~\citep{morris2020weisfeiler} which are more expressive than 1-WL GNNs. Note that all kernel versions usually perform much better than their neural counterparts. We use NNConv as our base GNN for QM9 and PNAConv for Alchemy and ZINC. For readout, we use Set2set output for QM9 following~\citep{gilmer2017neural} and sum pooling for other datasets. 
For all datasets we follow the dataset splits and report mean absolute error (MAE) as in~\citep{morris2020weisfeiler}. For more recent comparison with state-of-the-art models, we add PNA~\citep{corso2020principal} and DGN~\citep{beani2021directional}for ZINC and ALCHEMY datasets and LRBP-net~\citep{dupty2020neuralizing} for QM9.

\begin{table}[t]
    \LARGE
    \centering
	\resizebox{\columnwidth}{!}{
		\begin{tabular}{l|ccccc|ccc}
			\toprule
			& \multicolumn{5}{c|}{CSL} & \multicolumn{3}{c}{TRIANGLES}\\
			\midrule
			& mean & median & max  & min & std & Train & \multicolumn{2}{c}{Test}\\
			\cmidrule(lr){8-9}
			&  & &  & &   &  & orig & large  \\
			\midrule
			GIN & 10 & 10 & 10 & 10 & 0 & 90 & 47 & 18\\
		   \underline{GNN-IR with}   &&&&&&&& \\
		
			a) random $k$ nodes & 11.9 & 13.3 & 13.3 & 10 & 1.8  & 94.7 & 93.6 & 39.5\\
			b) without GRU & 90.7 & 96.7 & 96.7 & 70 & 16.6  & 97.1 & 97.2 & 41.2 \\
			c) sum aggr & 54.0 & 60.0 & 70.0 & 20 & 19.4 & 95.8 & 92.4 & 35.2\\
			GNN-IR & \textbf{98.7} & \textbf{100.0} & \textbf{100.0} & \textbf{96.67} & \textbf{1.8} & \textbf{99.4} & \textbf{99.3} & \textbf{51.1}\\
			\bottomrule
		\end{tabular}
	}
\caption{GNN-IR ablation models\label{table_ablation}}
\end{table}


Table~\ref{table3_zincalchemy} and~\ref{table3_qm9} shows that the improvement over standard baselines is consistent across the datasets. 
Specifically for ZINC10K and QM9, the improvement of GNN-IR is substantial over k-WL GNN variants. Note that it is non-trivial to compare the expressive power of GNN-IR in terms of k-WL models. Results show that GNN-IR outperforms neural versions of k-WL. This suggests that GNN-IR has at least better inductive bias for these problems than k-WL GNNs. But it is not clear which properties of the algorithm result in better inductive bias for these tasks. It would be interesting to understand the conditions under which GNN-IR and k-WL GNNs do well for different tasks. We leave this study for future work. 

\textbf{Runtime and ablation study:} We compare GNN-IR with k-WL GNNs in terms of their runtimes against accuracy. Table~\ref{table3_runtime} shows ratio of time per epoch of models w.r.t GINE (1-WL). There is only linear increase in runtime of GNN-IR with increasing $L,k$ and much better gains compared to  k-WL GNNs with lesser runtimes.

We also conducted ablation experiments to study the effectiveness of various components of GNN-IR. For this, we use the best models of GNN-IR for CSL and TRIANGLES dataset and consider randomly selecting top-k nodes instead of learning, replacing GRU in equation~\ref{eq:target_cell_2} with an MLP and using sum instead of max in aggregating the $k$ refined graphs. Table~\ref{table_ablation} shows that, compared to GNN, randomly selecting top-$k$ nodes helps in TRIANGLES dataset but does not help at all in CSL dataset; suggesting that learning top-k nodes is needed for breaking the symmetry in graph structures. Replacing GRU with MLP and using sum aggregator decreases the performance slightly but fare much better than GNN.

\textbf{Note:} Due to space constraints, we provide results on TUDataset benchmark in the Appendix; GNN-IR gives competitive performance against GNN-based methods although it does not give state-of-the-art performance when compared to all methods.

\section{Conclusion}\label{sec:conclusion}
In this work, we propose learning richer representations on graph structured data with \emph{individualization and refinement}, a technique followed by most practical isomorphism solvers. Our approach is computationally feasible and can adaptively select nodes to break the symmetry in GNN embeddings. Experimental evaluation shows that our model substantially outperforms other 1-WL and more expressive GNNs on several benchmark datasets. Future work includes understanding  the power and limitations of learning individualization and refinement functions for improving GNN models.


\bibliography{main_bibfile}

\begin{thebibliography}{57}
\providecommand{\natexlab}[1]{#1}

\bibitem[{Arvind et~al.(2020)Arvind, Fuhlbr{\"u}ck, K{\"o}bler, and
  Verbitsky}]{arvind2020weisfeiler}
Arvind, V.; Fuhlbr{\"u}ck, F.; K{\"o}bler, J.; and Verbitsky, O. 2020.
\newblock On Weisfeiler-Leman invariance: Subgraph counts and related graph
  properties.
\newblock \emph{Journal of Computer and System Sciences}, 113: 42--59.

\bibitem[{Barcel{\'o} et~al.(2019)Barcel{\'o}, Kostylev, Monet, P{\'e}rez,
  Reutter, and Silva}]{barcelo2019logical}
Barcel{\'o}, P.; Kostylev, E.~V.; Monet, M.; P{\'e}rez, J.; Reutter, J.; and
  Silva, J.~P. 2019.
\newblock The logical expressiveness of graph neural networks.
\newblock In \emph{International Conference on Learning Representations}.

\bibitem[{Beani et~al.(2021)Beani, Passaro, L{\'e}tourneau, Hamilton, Corso,
  and Li{\`o}}]{beani2021directional}
Beani, D.; Passaro, S.; L{\'e}tourneau, V.; Hamilton, W.; Corso, G.; and
  Li{\`o}, P. 2021.
\newblock Directional graph networks.
\newblock In \emph{International Conference on Machine Learning}, 748--758.
  PMLR.

\bibitem[{Bengio, Courville, and Vincent(2013)}]{bengio2013representation}
Bengio, Y.; Courville, A.; and Vincent, P. 2013.
\newblock Representation learning: A review and new perspectives.
\newblock \emph{IEEE transactions on pattern analysis and machine
  intelligence}, 35(8): 1798--1828.

\bibitem[{Bouritsas et~al.(2020)Bouritsas, Frasca, Zafeiriou, and
  Bronstein}]{bouritsas2020improving}
Bouritsas, G.; Frasca, F.; Zafeiriou, S.; and Bronstein, M.~M. 2020.
\newblock Improving graph neural network expressivity via subgraph isomorphism
  counting.
\newblock \emph{arXiv preprint arXiv:2006.09252}.

\bibitem[{Chen et~al.(2019{\natexlab{a}})Chen, Chen, Hsieh, Lee, Liao, Liao,
  Liu, Qiu, Sun, Tang et~al.}]{chen2019alchemy}
Chen, G.; Chen, P.; Hsieh, C.-Y.; Lee, C.-K.; Liao, B.; Liao, R.; Liu, W.; Qiu,
  J.; Sun, Q.; Tang, J.; et~al. 2019{\natexlab{a}}.
\newblock Alchemy: A quantum chemistry dataset for benchmarking ai models.
\newblock \emph{arXiv preprint arXiv:1906.09427}.

\bibitem[{Chen et~al.(2020)Chen, Chen, Villar, and Bruna}]{chen2020can}
Chen, Z.; Chen, L.; Villar, S.; and Bruna, J. 2020.
\newblock Can graph neural networks count substructures?
\newblock \emph{arXiv preprint arXiv:2002.04025}.

\bibitem[{Chen et~al.(2019{\natexlab{b}})Chen, Villar, Chen, and
  Bruna}]{chen2019equivalence}
Chen, Z.; Villar, S.; Chen, L.; and Bruna, J. 2019{\natexlab{b}}.
\newblock On the equivalence between graph isomorphism testing and function
  approximation with gnns.
\newblock \emph{arXiv preprint arXiv:1905.12560}.

\bibitem[{Chung et~al.(2014)Chung, Gulcehre, Cho, and
  Bengio}]{chung2014empirical}
Chung, J.; Gulcehre, C.; Cho, K.; and Bengio, Y. 2014.
\newblock Empirical evaluation of gated recurrent neural networks on sequence
  modeling.
\newblock \emph{arXiv preprint arXiv:1412.3555}.

\bibitem[{Corso et~al.(2020)Corso, Cavalleri, Beaini, Li{\`o}, and
  Veli{\v{c}}kovi{\'c}}]{corso2020principal}
Corso, G.; Cavalleri, L.; Beaini, D.; Li{\`o}, P.; and Veli{\v{c}}kovi{\'c}, P.
  2020.
\newblock Principal neighbourhood aggregation for graph nets.
\newblock \emph{arXiv preprint arXiv:2004.05718}.

\bibitem[{Darga et~al.(2004)Darga, Liffiton, Sakallah, and
  Markov}]{darga2004exploiting}
Darga, P.~T.; Liffiton, M.~H.; Sakallah, K.~A.; and Markov, I.~L. 2004.
\newblock Exploiting structure in symmetry detection for CNF.
\newblock In \emph{Proceedings of the 41st Annual Design Automation
  Conference}, 530--534.

\bibitem[{Dasoulas et~al.(2019)Dasoulas, Santos, Scaman, and
  Virmaux}]{dasoulas2019coloring}
Dasoulas, G.; Santos, L.~D.; Scaman, K.; and Virmaux, A. 2019.
\newblock Coloring graph neural networks for node disambiguation.
\newblock \emph{arXiv preprint arXiv:1912.06058}.

\bibitem[{Defferrard, Bresson, and
  Vandergheynst(2016)}]{defferrard2016convolutional}
Defferrard, M.; Bresson, X.; and Vandergheynst, P. 2016.
\newblock Convolutional neural networks on graphs with fast localized spectral
  filtering.
\newblock In \emph{Advances in neural information processing systems},
  3844--3852.

\bibitem[{Dehmamy, Barab{\'a}si, and Yu(2019)}]{dehmamy2019understanding}
Dehmamy, N.; Barab{\'a}si, A.-L.; and Yu, R. 2019.
\newblock Understanding the representation power of graph neural networks in
  learning graph topology.
\newblock \emph{arXiv preprint arXiv:1907.05008}.

\bibitem[{Du et~al.(2019)Du, Hou, P{\'o}czos, Salakhutdinov, Wang, and
  Xu}]{du2019graph}
Du, S.~S.; Hou, K.; P{\'o}czos, B.; Salakhutdinov, R.; Wang, R.; and Xu, K.
  2019.
\newblock Graph neural tangent kernel: Fusing graph neural networks with graph
  kernels.
\newblock \emph{arXiv preprint arXiv:1905.13192}.

\bibitem[{Dupty and Lee(2020)}]{dupty2020neuralizing}
Dupty, M.~H.; and Lee, W.~S. 2020.
\newblock Neuralizing Efficient Higher-order Belief Propagation.
\newblock \emph{arXiv preprint arXiv:2010.09283}.

\bibitem[{Dwivedi et~al.(2020)Dwivedi, Joshi, Laurent, Bengio, and
  Bresson}]{dwivedi2020benchmarking}
Dwivedi, V.~P.; Joshi, C.~K.; Laurent, T.; Bengio, Y.; and Bresson, X. 2020.
\newblock Benchmarking graph neural networks.
\newblock \emph{arXiv preprint arXiv:2003.00982}.

\bibitem[{Errica et~al.(2019)Errica, Podda, Bacciu, and
  Micheli}]{errica2019fair}
Errica, F.; Podda, M.; Bacciu, D.; and Micheli, A. 2019.
\newblock A fair comparison of graph neural networks for graph classification.
\newblock \emph{arXiv preprint arXiv:1912.09893}.

\bibitem[{Fey and Lenssen(2019)}]{Fey/Lenssen/2019}
Fey, M.; and Lenssen, J.~E. 2019.
\newblock Fast Graph Representation Learning with {PyTorch Geometric}.
\newblock In \emph{ICLR Workshop on Representation Learning on Graphs and
  Manifolds}.

\bibitem[{Gao and Ji(2019)}]{gao2019graph}
Gao, H.; and Ji, S. 2019.
\newblock Graph u-nets.
\newblock In \emph{international conference on machine learning}, 2083--2092.
  PMLR.

\bibitem[{Gilmer et~al.(2017)Gilmer, Schoenholz, Riley, Vinyals, and
  Dahl}]{gilmer2017neural}
Gilmer, J.; Schoenholz, S.~S.; Riley, P.~F.; Vinyals, O.; and Dahl, G.~E. 2017.
\newblock Neural message passing for quantum chemistry.
\newblock In \emph{Proceedings of the 34th International Conference on Machine
  Learning-Volume 70}, 1263--1272. JMLR. org.

\bibitem[{Hamilton, Ying, and
  Leskovec(2017{\natexlab{a}})}]{hamilton2017inductive}
Hamilton, W.; Ying, Z.; and Leskovec, J. 2017{\natexlab{a}}.
\newblock Inductive representation learning on large graphs.
\newblock In \emph{Advances in neural information processing systems},
  1024--1034.

\bibitem[{Hamilton, Ying, and
  Leskovec(2017{\natexlab{b}})}]{hamilton2017representation}
Hamilton, W.~L.; Ying, R.; and Leskovec, J. 2017{\natexlab{b}}.
\newblock Representation learning on graphs: Methods and applications.
\newblock \emph{arXiv preprint arXiv:1709.05584}.

\bibitem[{Jin, Barzilay, and Jaakkola(2018)}]{jin2018junction}
Jin, W.; Barzilay, R.; and Jaakkola, T. 2018.
\newblock Junction tree variational autoencoder for molecular graph generation.
\newblock In \emph{International Conference on Machine Learning}, 2323--2332.
  PMLR.

\bibitem[{Junttila and Kaski(2011)}]{junttila2011conflict}
Junttila, T.; and Kaski, P. 2011.
\newblock Conflict propagation and component recursion for canonical labeling.
\newblock In \emph{International Conference on Theory and Practice of
  Algorithms in (Computer) Systems}, 151--162. Springer.

\bibitem[{Kersting et~al.(2016)Kersting, Kriege, Morris, Mutzel, and
  Neumann}]{kersting2016benchmark}
Kersting, K.; Kriege, N.~M.; Morris, C.; Mutzel, P.; and Neumann, M. 2016.
\newblock Benchmark data sets for graph kernels, 2016.
\newblock \emph{URL http://graphkernels. cs. tu-dortmund. de}, 795.

\bibitem[{Kiefer et~al.(2020)Kiefer, Immerman, Schweitzer, and
  Grohe}]{kiefer2020power}
Kiefer, S.; Immerman, N.; Schweitzer, P.; and Grohe, M. 2020.
\newblock Power and limits of the Weisfeiler-Leman algorithm.
\newblock Technical report, Fachgruppe Informatik.

\bibitem[{Kipf and Welling(2016)}]{kipf2016semi}
Kipf, T.~N.; and Welling, M. 2016.
\newblock Semi-supervised classification with graph convolutional networks.
\newblock \emph{arXiv preprint arXiv:1609.02907}.

\bibitem[{Klicpera, Gro{\ss}, and
  G{\"u}nnemann(2020)}]{klicpera2020directional}
Klicpera, J.; Gro{\ss}, J.; and G{\"u}nnemann, S. 2020.
\newblock Directional message passing for molecular graphs.
\newblock \emph{arXiv preprint arXiv:2003.03123}.

\bibitem[{Knyazev, Taylor, and Amer(2019)}]{knyazev2019understanding}
Knyazev, B.; Taylor, G.~W.; and Amer, M.~R. 2019.
\newblock Understanding attention and generalization in graph neural networks.
\newblock \emph{arXiv preprint arXiv:1905.02850}.

\bibitem[{Lee, Lee, and Kang(2019)}]{lee2019self}
Lee, J.; Lee, I.; and Kang, J. 2019.
\newblock Self-attention graph pooling.
\newblock In \emph{International Conference on Machine Learning}, 3734--3743.
  PMLR.

\bibitem[{Li et~al.(2020)Li, Wang, Wang, and Leskovec}]{li2020distance}
Li, P.; Wang, Y.; Wang, H.; and Leskovec, J. 2020.
\newblock Distance Encoding--Design Provably More Powerful GNNs for Structural
  Representation Learning.
\newblock \emph{arXiv preprint arXiv:2009.00142}.

\bibitem[{Loukas(2019)}]{loukas2019graph}
Loukas, A. 2019.
\newblock What graph neural networks cannot learn: depth vs width.
\newblock In \emph{International Conference on Learning Representations}.

\bibitem[{Maron et~al.(2019)Maron, Ben-Hamu, Serviansky, and
  Lipman}]{maron2019provably}
Maron, H.; Ben-Hamu, H.; Serviansky, H.; and Lipman, Y. 2019.
\newblock Provably powerful graph networks.
\newblock \emph{arXiv preprint arXiv:1905.11136}.

\bibitem[{Maron et~al.(2018)Maron, Ben-Hamu, Shamir, and
  Lipman}]{maron2018invariant}
Maron, H.; Ben-Hamu, H.; Shamir, N.; and Lipman, Y. 2018.
\newblock Invariant and equivariant graph networks.
\newblock \emph{arXiv preprint arXiv:1812.09902}.

\bibitem[{McKay and Piperno(2014)}]{mckay2014practical}
McKay, B.~D.; and Piperno, A. 2014.
\newblock Practical graph isomorphism, II.
\newblock \emph{Journal of Symbolic Computation}, 60: 94--112.

\bibitem[{McKay et~al.(1981)}]{mckay1981practical}
McKay, B.~D.; et~al. 1981.
\newblock Practical graph isomorphism.

\bibitem[{Morris, Rattan, and Mutzel(2020)}]{morris2020weisfeiler}
Morris, C.; Rattan, G.; and Mutzel, P. 2020.
\newblock Weisfeiler and Leman go sparse: Towards scalable higher-order graph
  embeddings.
\newblock \emph{Advances in Neural Information Processing Systems}, 33.

\bibitem[{Morris et~al.(2019)Morris, Ritzert, Fey, Hamilton, Lenssen, Rattan,
  and Grohe}]{morris2019weisfeiler}
Morris, C.; Ritzert, M.; Fey, M.; Hamilton, W.~L.; Lenssen, J.~E.; Rattan, G.;
  and Grohe, M. 2019.
\newblock Weisfeiler and leman go neural: Higher-order graph neural networks.
\newblock In \emph{Proceedings of the AAAI Conference on Artificial
  Intelligence}, volume~33, 4602--4609.

\bibitem[{Murphy et~al.(2019)Murphy, Srinivasan, Rao, and
  Ribeiro}]{murphy2019relational}
Murphy, R.; Srinivasan, B.; Rao, V.; and Ribeiro, B. 2019.
\newblock Relational pooling for graph representations.
\newblock In \emph{International Conference on Machine Learning}, 4663--4673.
  PMLR.

\bibitem[{Nikolentzos and Vazirgiannis(2020)}]{nikolentzos2020random}
Nikolentzos, G.; and Vazirgiannis, M. 2020.
\newblock Random Walk Graph Neural Networks.
\newblock \emph{Advances in Neural Information Processing Systems}, 33:
  16211--16222.

\bibitem[{Qi et~al.(2017)Qi, Su, Mo, and Guibas}]{qi2017pointnet}
Qi, C.~R.; Su, H.; Mo, K.; and Guibas, L.~J. 2017.
\newblock Pointnet: Deep learning on point sets for 3d classification and
  segmentation.
\newblock In \emph{Proceedings of the IEEE conference on computer vision and
  pattern recognition}, 652--660.

\bibitem[{Ramakrishnan et~al.(2014)Ramakrishnan, Dral, Rupp, and
  Von~Lilienfeld}]{ramakrishnan2014quantum}
Ramakrishnan, R.; Dral, P.~O.; Rupp, M.; and Von~Lilienfeld, O.~A. 2014.
\newblock Quantum chemistry structures and properties of 134 kilo molecules.
\newblock \emph{Scientific data}, 1: 140022.

\bibitem[{Ruddigkeit et~al.(2012)Ruddigkeit, Van~Deursen, Blum, and
  Reymond}]{ruddigkeit2012enumeration}
Ruddigkeit, L.; Van~Deursen, R.; Blum, L.~C.; and Reymond, J.-L. 2012.
\newblock Enumeration of 166 billion organic small molecules in the chemical
  universe database GDB-17.
\newblock \emph{Journal of chemical information and modeling}, 52(11):
  2864--2875.

\bibitem[{Sato, Yamada, and Kashima(2020)}]{sato2020random}
Sato, R.; Yamada, M.; and Kashima, H. 2020.
\newblock Random features strengthen graph neural networks.
\newblock \emph{arXiv preprint arXiv:2002.03155}.

\bibitem[{Sch{\"u}tt et~al.(2018)Sch{\"u}tt, Sauceda, Kindermans, Tkatchenko,
  and M{\"u}ller}]{schutt2018schnet}
Sch{\"u}tt, K.~T.; Sauceda, H.~E.; Kindermans, P.-J.; Tkatchenko, A.; and
  M{\"u}ller, K.-R. 2018.
\newblock SchNet--A deep learning architecture for molecules and materials.
\newblock \emph{The Journal of Chemical Physics}, 148(24): 241722.

\bibitem[{Shervashidze et~al.(2011)Shervashidze, Schweitzer, Van~Leeuwen,
  Mehlhorn, and Borgwardt}]{shervashidze2011weisfeiler}
Shervashidze, N.; Schweitzer, P.; Van~Leeuwen, E.~J.; Mehlhorn, K.; and
  Borgwardt, K.~M. 2011.
\newblock Weisfeiler-lehman graph kernels.
\newblock \emph{Journal of Machine Learning Research}, 12(9).

\bibitem[{Srinivasan and Ribeiro(2019)}]{srinivasan2019equivalence}
Srinivasan, B.; and Ribeiro, B. 2019.
\newblock On the Equivalence between Positional Node Embeddings and Structural
  Graph Representations.
\newblock In \emph{International Conference on Learning Representations}.

\bibitem[{Unke and Meuwly(2019)}]{unke2019physnet}
Unke, O.~T.; and Meuwly, M. 2019.
\newblock PhysNet: A neural network for predicting energies, forces, dipole
  moments, and partial charges.
\newblock \emph{Journal of chemical theory and computation}, 15(6): 3678--3693.

\bibitem[{Vignac, Loukas, and Frossard(2020)}]{vignac2020building}
Vignac, C.; Loukas, A.; and Frossard, P. 2020.
\newblock Building powerful and equivariant graph neural networks with
  structural message-passing.
\newblock \emph{arXiv e-prints}, arXiv--2006.

\bibitem[{Vinyals, Bengio, and Kudlur(2015)}]{vinyals2015order}
Vinyals, O.; Bengio, S.; and Kudlur, M. 2015.
\newblock Order matters: Sequence to sequence for sets.
\newblock \emph{arXiv preprint arXiv:1511.06391}.

\bibitem[{Wu et~al.(2020)Wu, Pan, Chen, Long, Zhang, and
  Philip}]{wu2020comprehensive}
Wu, Z.; Pan, S.; Chen, F.; Long, G.; Zhang, C.; and Philip, S.~Y. 2020.
\newblock A comprehensive survey on graph neural networks.
\newblock \emph{IEEE Transactions on Neural Networks and Learning Systems}.

\bibitem[{Xu et~al.(2018)Xu, Hu, Leskovec, and Jegelka}]{xu2018powerful}
Xu, K.; Hu, W.; Leskovec, J.; and Jegelka, S. 2018.
\newblock How powerful are graph neural networks?
\newblock \emph{arXiv preprint arXiv:1810.00826}.

\bibitem[{Yanardag and Vishwanathan(2015)}]{yanardag2015deep}
Yanardag, P.; and Vishwanathan, S. 2015.
\newblock Deep graph kernels.
\newblock In \emph{Proceedings of the 21th ACM SIGKDD international conference
  on knowledge discovery and data mining}, 1365--1374.

\bibitem[{Ying et~al.(2018)Ying, You, Morris, Ren, Hamilton, and
  Leskovec}]{ying2018hierarchical}
Ying, R.; You, J.; Morris, C.; Ren, X.; Hamilton, W.~L.; and Leskovec, J. 2018.
\newblock Hierarchical graph representation learning with differentiable
  pooling.
\newblock \emph{arXiv preprint arXiv:1806.08804}.

\bibitem[{Zaheer et~al.(2017)Zaheer, Kottur, Ravanbakhsh, Poczos,
  Salakhutdinov, and Smola}]{zaheer2017deep}
Zaheer, M.; Kottur, S.; Ravanbakhsh, S.; Poczos, B.; Salakhutdinov, R.; and
  Smola, A. 2017.
\newblock Deep sets.
\newblock \emph{arXiv preprint arXiv:1703.06114}.

\bibitem[{Zhang et~al.(2018)Zhang, Cui, Neumann, and Chen}]{zhang2018end}
Zhang, M.; Cui, Z.; Neumann, M.; and Chen, Y. 2018.
\newblock An end-to-end deep learning architecture for graph classification.
\newblock In \emph{Proceedings of the AAAI Conference on Artificial
  Intelligence}, volume~32.

\end{thebibliography}


\newpage
\onecolumn
\appendix
\section{Appendix}
In this section, we first provide proofs of the propositions. Thereafter, we expand our Experiments section by including additional set of experiments. Specifically, we provide results on TUDataset benchmark and additional details on experiments and datasets.

\subsection{Proofs}

\textbf{Permutation invariance: }

\noindent\textbf{Lemma 1. } \emph{If the input to a GNN-IR round is an orbit-partitioned coloring, then the output will also be an orbit-partitioned coloring.}\\

\noindent\textit{Proof:} Each round of GNN-IR consists of selection of $k$ nodes, individualization-refinement of $k$ nodes to generate $k$ refined colorings and thereafter, aggregation of $k$ colorings into a single coloring. Firstly, since input is an orbit-partition, the node selection stage is permutation-invariant. The $k$ refined colorings generated after individualization-refinement stage are all orbit-partitions. This is because any refined sub-partition of an orbit-partition is also an orbit-partition. As a proof by contradiction for this: assume the sub-partition is not an orbit-partition. Then this implies there exist two vertices in a color-cell of the sub-partition which cannot be mapped to each other via an automorphism. From the definition of color-refinement, all color cells of sub-partition are subsets of some orbit-cell in the initial orbit-partition which implies there exists an automorphism between the two vertices which is a contradiction of the assumption. Hence, the refined colorings are orbit-partitions. 

Now, the $k$ refined colorings of the initial orbit-partition are mapped to a single coloring by the aggregation operation. Given that all the $k$ refined colorings are finer than the initial coloring, we need to show that any node-wise aggregation will either be finer or at least equal to the initial orbit-partition in terms of the number and size of the color cells, though the embeddings may be transformed. Equivalently, if two vertices have different colors before IR step, then they should have different colors after IR step.

Consider the node-wise aggregation operation. It takes $k$ colorings as input and for each node, injectively maps the set of $k$ colors/embeddings to a new color. If two vertices have different colors in at least one of the $k$ colorings, then they will be mapped to different colors except in one case. The only case when two vertices $v_a$ and $v_b$ with different colors in at least one of the $k$ colorings, are mapped to the same color is when the  refinements are rotations of each other. For simplicity, consider aggregating $k=2$ refinements; similar argument follows for higher values of $k$. Let $v_i$ and $v_j$ be the individualizing vertices for generating the 2 refinements and $\pi_{v_i}^l(v)$ be the color of $v$ in the refined coloring after individualizing $v_i$. If the two refined colorings are rotations of each other, then $\pi_{v_i}^l(v_a) = \pi_{v_j}^l(v_b)$ and $\pi_{v_j}^l(v_a) = \pi_{v_i}^l(v_b)$ is true for some vertices $v_a$ and $v_b $. Then, when we aggregate nodes across $k$ colorings as a set of colors, the set becomes same for both $v_a$ and $v_b$ i.e., $\{\pi_{v_i}^l(v_a), \pi_{v_j}^l(v_a)\} = \{\pi_{v_i}^l(v_b), \pi_{v_j}^l(v_b)\}$. Below, we show that such a condition cannot occur for any pair of vertices with different colors in the initial orbit-partition. 

For this, consider two vertices $v_a$ and $v_b$ with different colors in the initial orbit-partition i.e. $\pi^l(v_a) \neq \pi^l(v_b)$ where $\pi^l$ is the coloring before layer $l$ of GNN-IR. For clarity, let $x$ and $y$ be the colors of $v_a$ and $v_b$ respectively with $x\neq y$, in the initial orbit-partition. Now, if the individualizing vertices $v_i$ and $v_j$ belong to different color-cells, then the refinements induced will be different. This can be shown from the procedure of vertex-refinement~\cite{shervashidze2011weisfeiler}. Therefore, the new colors of $v_a$ and $v_b$ will be different in both the refinements i.e. colorings due to $v_i$ will be $x', y'$ and colorings due to $v_j$ will be $x'', y''$. Note that $x'\neq y''$ and  $y'\neq x''$, since after vertex-refinement, the new color of a vertex depends both on its previous color and color of individualizing vertex. In this case, we have different colors for all four vertices i.e., $\pi^l(v_i)\neq\pi^l(v_j)\neq\pi^l(v_a)\neq\pi^l(v_b)$ which implies $x'\neq x'' \neq y'\neq y''$. Therefore, the node-wise aggregated sets are $\{x',x''\} \neq \{y',y''\}$ and hence, $\pi^{l+1}(v_a) \neq \pi^{l+1}(v_b)$ after aggregation.

If the individualizing vertices $v_i$ and $v_j$ belong to same color-cells, then the refinements induced will be same. But since their initial colors were different, the node-wise aggregated set of colors would be different for each vertex. Equivalently, if the colors of $v_a, v_b$ due to individualizing $v_i$ are $x', y'$ respectively, then the colors due to $v_j$ will also be $x', y'$. In this case, the node-wise aggregated sets are $\{x',x'\} \neq \{y',y'\}$ and hence, $\pi^{l+1}(v_a) \neq \pi^{l+1}(v_b)$ after aggregation.
	
Hence, the aggregation operation generates coloring which is also an orbit partition. \hfill \qedsymbol\\

\noindent \textbf{Expressive power of GNN-IR: }
We now proceed to the proof of Proposition 1 which states that GNN-IR is more expressive than GNN. For the proof, we need to know the conditions under which the aggregation procedure of GNN-IR as described in Algorithm~\ref{alg:GNN-IR} maps two sets of $k$ colorings to different colorings. The following Lemma shows one of the conditions under which the aggregation can map refinements of two non-isomorphic graphs to distinct aggregated colorings.

\noindent\textbf{Lemma 2. } \emph{Assume we use we use universal set approximators in aggregation function to merge $k$ refined colorings (colored graphs) in Algorithm~\ref{alg:GNN-IR}. Consider two sets of $k$ colorings $\Pi_a$ and $\Pi_b$ such that $\Pi_a = \{\pi_{a_1}, \pi_{a_2},\dots,\pi_{a_k}\}$ and $\Pi_b = \{\pi_{b_1}, \pi_{b_2},\dots,\pi_{b_k}\}$. If $\Pi_a \cap \Pi_b = \varnothing$ i.e.$\Pi_a$ and $\Pi_b$ do not share any coloring out of $k$ colorings, then, irrespective of the node embeddings, the aggregation function of GNN-IR maps $\Pi_a$ and $\Pi_b$ to different colorings.}\\

\noindent\textit{Proof:} 
With the assumption of universal set approximators for AGG in line 15 and 18 of Algorithm~\ref{alg:GNN-IR}, the aggregation function first maps each $i^{th}$ coloring to a unique $\tilde{h}_G^{l^i}$ embedding. Note that, we then concatenate $\tilde{h}_G^{l^i}$ with each of the node embeddings of the $i^{th}$ coloring. We then use another set function approximator to reduce $k$ node embeddings to one. For $\Pi_a$ and $\Pi_b$, since $\Pi_a \cap \Pi_b = \varnothing$, the corresponding set of $\tilde{h}_G^{l^i}$'s will be different and consequently,  irrespective of node embeddings $\tilde{h}_v^{l^i}$, every vertex across $k$ colorings,  will be mapped to different embedding after the concatenation $[\tilde{h}_v^{l^i},\tilde{h}_G^{l^i}]$. Therefore, with the set aggregation of the $k$ concatenated node embeddings, 
$\Pi_a$ and $\Pi_b$ will be mapped to different colorings. \hfill \qedsymbol\\

Below, we give the extended proof of Proposition 1.

\noindent\textbf{Proposition 1. } \emph{Assume we use universal set approximators in GNN-IR for target-cell selection, refinement and aggregation steps. GNN-IR is more expressive than all 1-WL equivalent GNNs i.e., GNN-IR can distinguish all graphs distinguishable by GNN and there exist graphs non-distinguishable by GNN which can be distinguished by GNN-IR.}\\

\noindent\textit{Proof:} 
Consider the procedure of generating the final embedding of GNN-IR used for graph prediction. After every step of GNN-IR, we pool the node embeddings for readout. Then we concatenate the pooled embeddings of each layer and feed it to an MLP for the final prediction. If $h_G^l$ is the pooled graph embedding after layer $l$, then the concatenated embedding will be $[h_G^0, h_G^1, \dots, h_G^L]$ for GNN-IR with $L$ layers. Clearly, this concatenated embedding must be different for all the graphs distinguishable by GNN-IR.

First, consider $\mathcal{G}^{1WL}$ as the set of graphs such that the equitable coloring induced after 1-WL/GNN refinement on any $G \in \mathcal{G}^{1WL}$ uniquely identifies $G$.  In other words, $\mathcal{G}^{1WL}$ is the set of graphs which are distinguishable by any 1-WL equivalent GNN. Note that if an equitable coloring $\pi$ uniquely identifies a graph, then its color-cells form the vertex orbits of the automorphism group of the graph~\cite{kiefer2020power}. Therefore, the equitable coloring induced by GNN refinement on any $G \in \mathcal{G}^{1WL}$ also forms an orbit partition and GNN-IR with any width $k$ preserves its permutation-invariance. 
Therefore, the final GNN-IR embedding $[h_G^0, h_G^1, \dots, h_G^L]$ will be permutation-invariant regardless of width $k$ and length $L$. Now, since $G \in \mathcal{G}^{1WL}$ is distinguishable by GNN, $h_G^0$ will be unique for $G$ and will be distinguishable by GNN-IR. 

\begin{figure}[h]
	\centering
	\begin{subfigure}{.5\textwidth}
		\centering
		\includegraphics[width=0.9\linewidth,height=0.41\columnwidth]{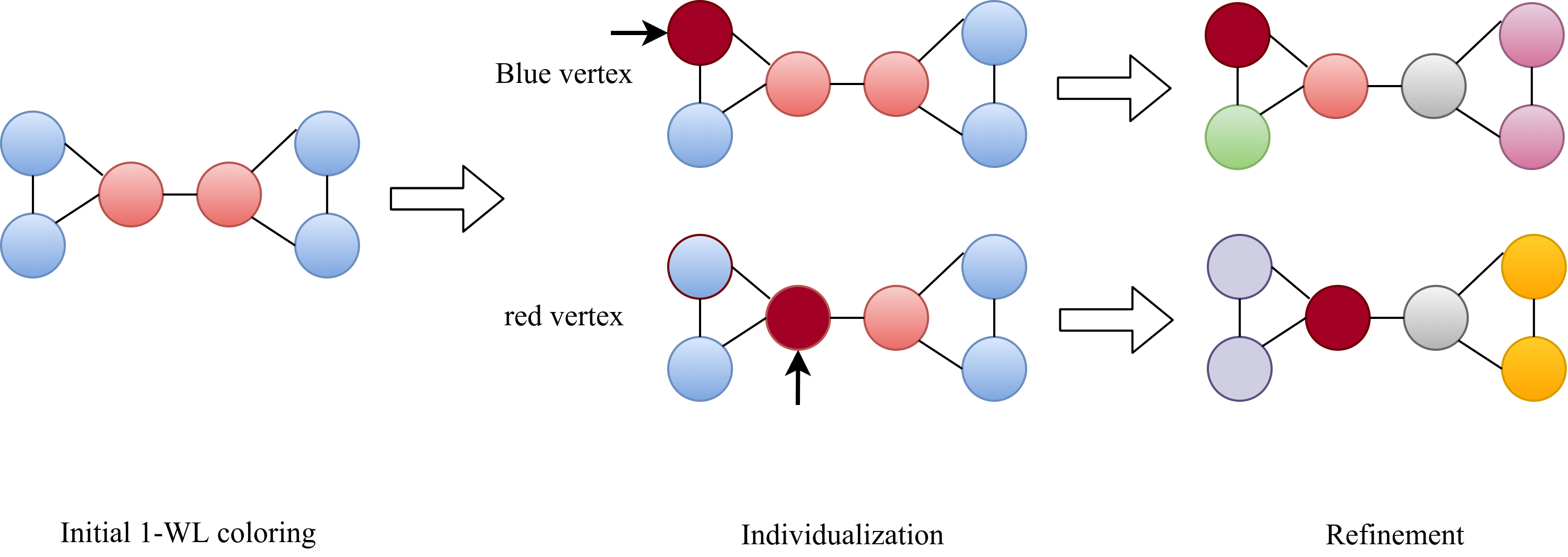}
		\caption{Graph A}
		\label{fig:sub1}
	\end{subfigure}%
	\begin{subfigure}{.5\textwidth}
		\centering
		\includegraphics[width=0.9\linewidth,height=0.41\columnwidth]{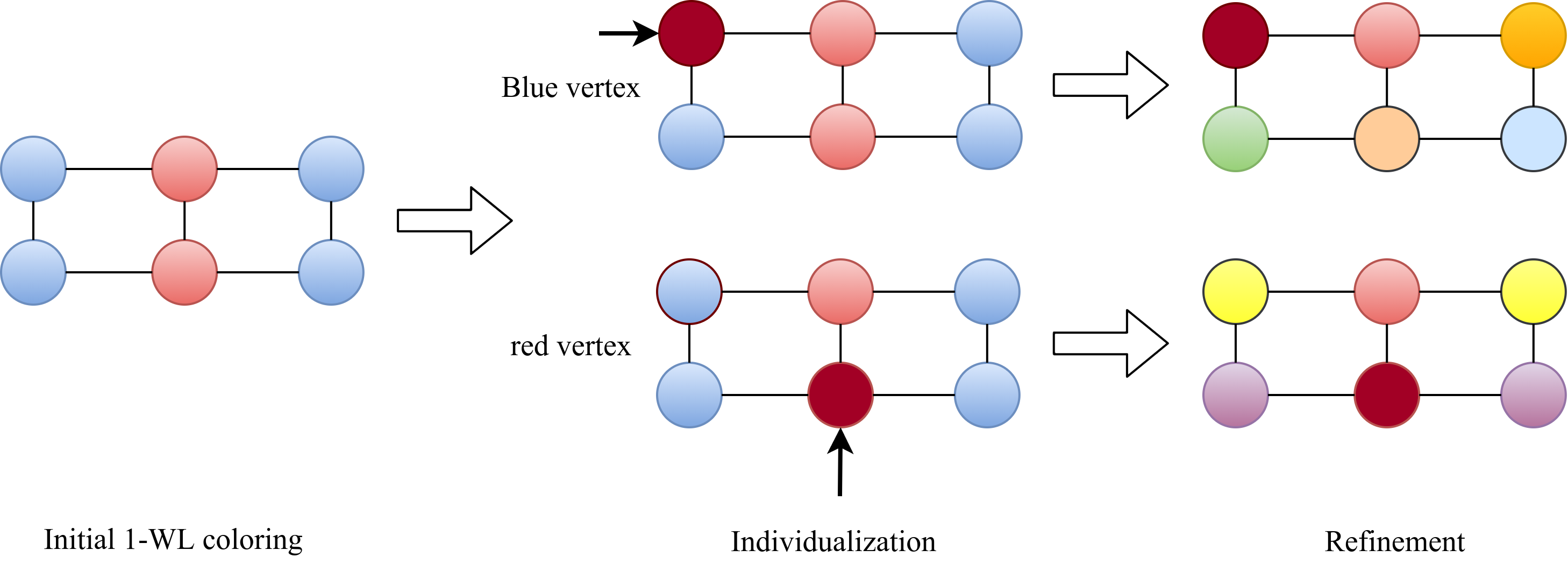}
		\caption{Graph B}
		\label{fig:sub2}
	\end{subfigure}
	\caption{Two non-isomorphic graphs inducing different refinements for both color classes. Note that in both Graphs A and B, the initial 1-WL coloring forms orbit partition and hence refinements induced are same for all vertices of either blue or red vertex class in the graph.}
	\label{fig:proof-illustration}
\end{figure}

Next, we need to show that there exist graphs which are not distinguishable by any 1-WL equivalent GNN but are distinguishable by GNN-IR. Graphs A and B in Figure~\ref{fig:proof-illustration} are two such graphs which are not distinguishble by GNN and hence, $h_G^0$ will be same for these two graphs. It can be shown that initial 1-WL refinement of graphs A and B generates orbit partitioned colorings. Such a refinement will effectively partition the graphs into two cells of orbits i.e. blue and red colored vertices. Since, the initial 1-WL refinement of both the graphs generates orbit-partitioned colorings, 
by Lemma 1, GNN-IR preserves permutation-invariance 
for any number of IR steps.

As illustrated in Figure~\ref{fig:proof-illustration}, the individualization of either of the blue or red vertices induces different refined colorings for graphs A and B. Since the number of vertices are 6, any value of $k<=6$ will result in distinct multiset of colorings $\Pi_A$ and $\Pi_B$ for graphs A and B respectively, such that $\Pi_A \cap \Pi_B = \varnothing$. With Lemma 2., the aggregation function maps the two graphs to distinct colorings which can be pooled to get $h_G^l$. Hence, in the final embedding $[h_G^0, h_G^1, \dots, h_G^L]$, $h_G^l$ for any $l>0$, will be different for the two graphs.
Therefore, these graphs can be distinguished by GNN-IR.\hfill \qedsymbol\\ 

\subsection{Experiments}\label{subsec:experiments}
\subsubsection{Graph classification on TUDataset bechmark}\label{subsubsec:TUdataset}

We evaluate GNN-IR on six datasets from the publicly available TUDataset bechmark, the graph classification benchmark suite~\citep{kersting2016benchmark, yanardag2015deep}. The datasets are broadly in domains of social media (IMDB-BINARY, IMDB-MULTI, COLLAB) and chemical science (NCI1, PROTEINS, MUTAG). We use GIN as our base GNN convolution operator and sum/mean pooling for readout.\\

\textbf{Experimental setup} 

The evaluation procedures on TUDataset vary considerably in the recent literature on GNNs. This has given rise to different numbers for the same models. One evaluation setup~\citep{xu2018powerful} is to split the datasets into training and validation sets in the ratio of 90:10. Thereafter, take the mean of the validation curves across 10 folds and report mean/std of the best epoch of the mean validation curve. This evaluation protocol is non-standard but is used because of the small size of datasets. However, recently~\citet{errica2019fair} did a fair comparison of the prominent GNN models. They performed 10-fold standard 90:10 train-test split and report the average accuracy on the test set while model selection is done on a separate 10\% split on the training set. We adopt this standard method of evaluation and report the accuracies in 10-fold experiment as described in~\citet{errica2019fair} and~\citet{nikolentzos2020random}.

Note that we are only comparing with neural GNN models and do not claim state-of-the-art score on these datasets. Kernel versions like~\cite{du2019graph,morris2020weisfeiler} usually perform better than their neural counterparts on these particular datasets. In fact, the higher-order Local $k$-WL models in kernel versions have the state-of-the-art score for most of the TUDatasets. ~\cite{morris2020weisfeiler} show results of only kernel versions of "Local $k$-WL" models on TUDatasets and not neural versions. Nonetheless, since we are comparing with higher-order $k$-WL GNNs in our experiments, we report results of 3-WL-GNN with this setup as well.

For our experiments, we compare the proposed GNN-IR model against the following GNN baselines : DGCNN~\citep{zhang2018end}, DiffPool~\citep{ying2018hierarchical}, GIN~\citep{xu2018powerful}, GraphSAGE~\citep{hamilton2017inductive} and RWNN~\citep{nikolentzos2020random} and the higher-order 3-WL-GNN~\citep{morris2019weisfeiler}. Results for these baselines are as reported in~\cite{nikolentzos2020random} and we use the code of 1-2-3-GNN~\citep{morris2019weisfeiler} to generate results for 3-WL-GNN with this setup.

\begin{table}[h]
	\centering
	\begin{minipage}{\linewidth}
		\begin{center}
			\begin{scriptsize}
				\resizebox{0.95\textwidth}{!}{
					\begin{tabular}{lcccccc}
						\toprule
						Data set & IMDB-B & IMDB-M & COLLAB  & NCI1 & PROTEINS & MUTAG \\
						\midrule
						DGCNN & 69.2{\tiny$\pm3.0$} & 45.6{\tiny$\pm3.4$}  & 71.2{\tiny$\pm1.9$} & 76.4{\tiny$\pm1.7$} & 72.9{\tiny$\pm3.5$} & 84.0{\tiny$\pm6.7$}\\
						Diffpool & 68.4{\tiny$\pm3.3$} & 45.6{\tiny$\pm3.4$} &  68.9{\tiny$\pm2.0$} & 76.9{\tiny$\pm1.9$} & 73.7{\tiny$\pm3.5$} & 79.8{\tiny$\pm7.1$}\\
						GIN$^*$ & 71.2{\tiny$\pm3.9$} & 48.5{\tiny$\pm3.3$} &  75.6{\tiny$\pm2.3$} & 80.0{\tiny$\pm1.4$} & 73.3{\tiny$\pm4.0$} & 84.7{\tiny$\pm6.7$} \\
						GRAPHSAGE & 68.8{\tiny$\pm4.5$} & 47.6{\tiny$\pm3.5$} & 73.9{\tiny$\pm1.7$} & 76.0{\tiny$\pm1.8$} & 73.0{\tiny$\pm4.5$} & 83.6{\tiny$\pm9.6$}\\
						RWNN & 70.8{\tiny$\pm4.8$} & 48.8{\tiny$\pm2.9$}& 71.9{\tiny$\pm2.5$} & 73.9{\tiny$\pm1.3$} & 74.7{\tiny$\pm3.3$}& \bf{89.2}{\tiny$\pm4.3$}\\
						3-WL-GNN & 71.5{\tiny$\pm3.6$} & \bf{50.4}{\tiny$\pm4.5$} & OOM &  75.7{\tiny$\pm3.4$} & 74.9{\tiny$\pm5.4$} & 86.2{\tiny$\pm7.0$}\\
						\midrule
						GNN-IR & \bf{73.7}{\tiny$\pm4.3$} & 50.1{\tiny$\pm3.8$} &  \bf{77.7}{\tiny$\pm3.1$} & \bf{80.7}{\tiny$\pm2.2$} &  \bf{75.5}{\tiny$\pm2.2$} & $88.7${\tiny$\pm6.9$}\\
						\bottomrule
					\end{tabular}
				}
			\end{scriptsize}
		\end{center}
			\caption{\footnotesize Graph classification on TUDataset. Our convolution operator is GINconv. OOM is Out of Memory.}
	\label{tab:Tudatasets_1}
	\end{minipage}
\end{table}

\begin{table}[h]
	\centering
	\begin{minipage}{0.95\linewidth}

		\begin{center}
			\begin{scriptsize}
				\resizebox{0.55\columnwidth}{!}{
					\begin{tabular}{lccc}
						\toprule
						Data set & IMDB-B & IMDB-M & COLLAB  \\
						\midrule
						DGCNN & 53.3{\tiny$\pm 5.0$} & 38.6{\tiny$\pm2.2$} & 57.4{\tiny$\pm1.9$} \\
						Diffpool & 68.3{\tiny$\pm6.1$} & 45.1{\tiny$\pm3.2$} & 67.7{\tiny$\pm1.9$}\\
						GIN & 66.8{\tiny$\pm3.9$} & 42.2{\tiny$\pm4.6$} & \bf{75.9}{\tiny$\pm1.9$} \\
						GRAPHSAGE & 69.9{\tiny$\pm4.6$} & 47.2{\tiny$\pm 3.6$} & 71.6{\tiny$\pm1.5$} \\
						\midrule
						GNN-IR & \bf{71.9}{\tiny$\pm4.5$} & \bf{48.4}{\tiny$\pm5.5$}&   71.8{\tiny$\pm2.1$} \\
						\bottomrule
					\end{tabular}
				}
			\end{scriptsize}
		\end{center}
			\caption{\footnotesize Graph classification on social media datasets without node-degree as features.}
	\label{tab:Tudatasets_2}
	\end{minipage}
	\vskip 0.15in
\end{table}

\textbf{Without node-degree features}

The social media datasets do not come with node attributes and are initialized with node-degrees as node features. In principle, since node-degrees can be computed by 1-WL message passing, the presence of node-degree features should not affect the performance of GNN models. However, in practice GNN models perform poorly without node-degree features. This was shown in the results of~\cite{errica2019fair}. In order to assess the GNN-IR's robustness to the presence of node-degree features, we report additional results without node-degree features as well. In this case, we initialize node features with constant values. 

\textbf{Results}

Table~\ref{tab:Tudatasets_1} shows the accuracy of the GNN models on graph classification in TUDataset. Clearly, GNN-IR scores are competitive in all the datasets. GNN-IR outperforms higher-order 3-WL-GNN by significant margin in IMDB-B, NCI1 datasets while in other datasets increase of 1-2\% points can be seen. Compared to all other baselines, GNN-IR achieves the best score in 4 out of 6 datasets and is close to the best performing in other 2 datasets. Note the results in comparison to GIN, which is the base convolution operator used in GNN-IR. GNN-IR's clear edge over GIN shows that the improvement is coming from the \emph{individualization and refinement} mechanism of GNN-IR.

Furthermore, the difference is clearer in Table~\ref{tab:Tudatasets_2}, which shows the accuracy of the models on graphs without node-degree features. With the exception of COLLAB dataset, accuracy of GNN-IR does not decrease significantly in both IMDB datasets. This suggests that the GNN-IR can capture structural information from the graphs significantly better than the 1-WL equivalent GNN models. Since, GNN-IR works by breaking the symmetry between nodes, it is more robust to the absence of node-degree features.

\begin{figure}[t]
	\centering     
		\resizebox{0.85\textwidth}{!}{
	\begin{minipage}{0.7\textwidth}	
		\begin{subfigure}[t]{0.5\textwidth}
			\includegraphics[width=0.85\textwidth]{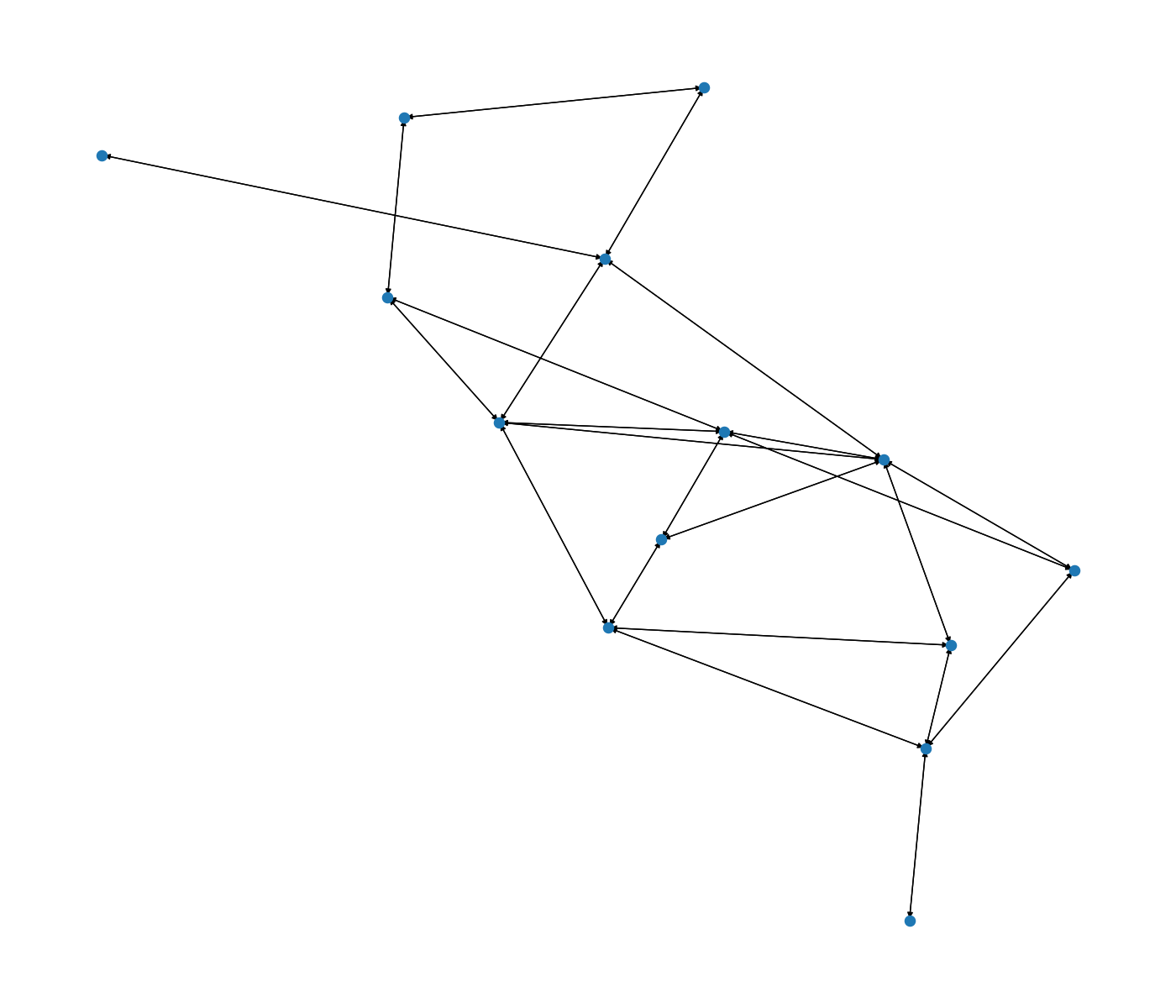}
			\label{fig:traingles_a}
			\caption{Test Orig, $Y$ = 6}
		\end{subfigure}\hfil
		\begin{subfigure}[t]{0.5\textwidth}
			\includegraphics[width=0.85\textwidth]{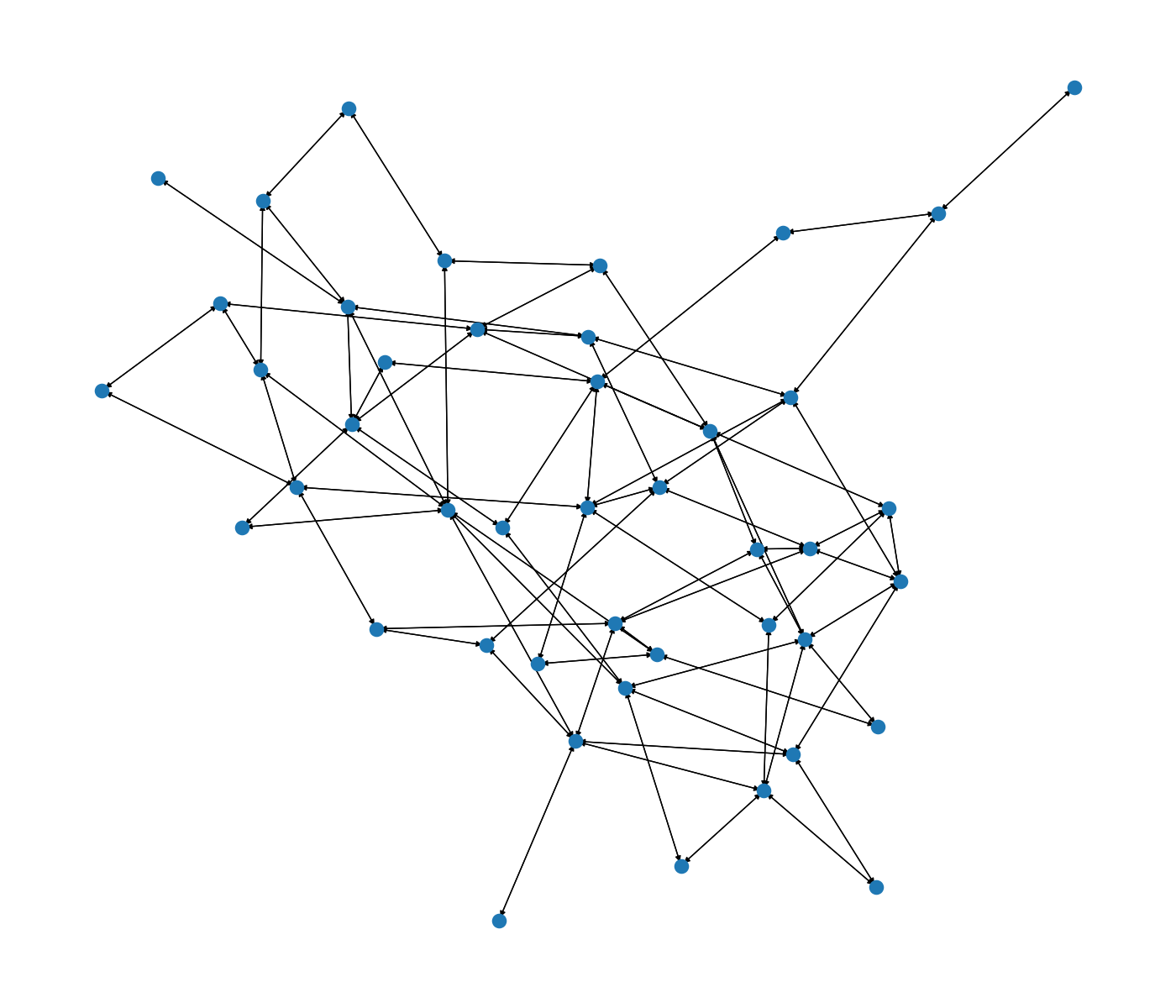}
			\label{fig:traingles_b}
			\caption{Test Large, $Y$ = 6}
		\end{subfigure}\hfil
		\caption{\footnotesize Graphs with $Y=6$ triangles from the \textsc{TRIANGLES} dataset.}
		\label{fig:triangles}
	\end{minipage}\hfil}
\end{figure}

\subsubsection{Experimental details}\label{subsubsec:exp_details}
In this section, we give further details of experiments conducted. The code was written in Pytorch Geometric~\citep{Fey/Lenssen/2019} library and the experiments were performed on a GeForce RTX 2080Ti GPU card. Hyperparameter tuning was done on a separate validation set formed from the training set on all the datasets. For a fair comparison, we used implementations of~\citet{morris2020weisfeiler} for ZINC10K, ALCHEMY10K, TUDataset bechmark and Pytorch Geometric example implementations for the rest of the datasets. 

For all the datasets, our message passing model consists of a convolution operator for message function and a GRU for update following the MPNN implementation of~\citet{gilmer2017neural}. The specific convolution operator used is given in Table~\ref{tab:tud_datasets_hyperparams} for each dataset. Additionally, we share the Conv-GRU parameters across the IR layers as this architecture scales well for deeper layers without increasing the parameter load and does not lose performance as well. Also, it can be said that the improvement shown in model's performance across datasets is not because of using more parameters. In each IR layer, GNN is run for $3$ steps. In the set aggregators, we use 1-hidden layer MLP before sum-pooling the vectors of the set. And finally, in datasets with edge features, we treat edges as variables and readout from both node and edge variables before the final fully connected layer.

We use cross entropy and mean absolute error (mae) loss functions for graph classification and regression respectively. We report dataset statistics and all the hyperparameters which were used for the results in Table~\ref{tab:tud_datasets_hyperparams}. These hyperparameters were chosen with a separate validation set for each dataset. Below we describe details of all datasets and the splits used in the experiments.

\begin{table*}[h]
	\centering
	\caption{Dataset statistics and hyperparameters used for all the datasets}
	\label{tab:tud_datasets_hyperparams}
	\resizebox{\textwidth}{!}{
		\begin{tabular}{lccccccccccc}
			Dataset & TRIANGLES & CSL & ZINC10K & ALCHEMY10K & QM9 & COLLAB & IMDB-B & IMDB-M & NCI1 & PROTEINS & MUTAG \\
			\midrule
			
			\#graphs & $45000$ & $150$ & $12000$ & $12000$ & $130831$ & $5000$ & $1000$ & $1500$ & $4,110$ & $1113$ & $188$ \\
			Node feat & Yes & No & Yes & Yes & Yes & No & No & No & Yes & Yes & Yes \\
			Edge feat & No & No & Yes & Yes & Yes & No & No & No & No & No & No \\
			\midrule
			
			batch size & $60$ & $16$ & $128$ & $128$ & $64$ & $64$ & $64$ & $64$ &$64$ & $64$ & $64$\\
			
			hidden layer size  & $64$ & $64$  & $75$ & $75$ & $64$ & $64$ & $64$ & $64$& $64$ & $64$ & $64$\\
			
			epochs & $300$ & $300$ & $200$ & $200$ & $200$& $200$ & $200$ & $200$  & $150$ & $150$ & $50$ \\
			
			start lr & $10^{-3}$ & $10^{-3}$ & $10^{-3}$ & $10^{-3}$ & $10^{-3}$ & $10^{-3}$ & $10^{-3}$ & $10^{-3}$ & $10^{-3}$ & $10^{-3}$ & $10^{-3}$\\
			
			decay rate  & $0.5$ &  $0.5$ &  $0.5$ &  $0.5$ &  $0.7$ &  $0.7$ & $0.7$ &  $0.7$ & $0.7$ &  $0.7$ &  $0.7$ \\
			
			decay steps & $-$ & $-$ & $-$ & $-$ & $-$ & $50$  & $50$ &  $50$ & $-$ & $-$ & $-$ \\
			
			patience & $15$ & $15$ & $15$ & $15$ & $5$ & $-$ & $-$ & $-$ & $15$ & $15$ & $10$ \\
			
			\#IR layers $L$ & $-$ & $-$ & $2$ & $2$ & $2$ & $3$ & $3$ & $2$ & $2$ & $2$ & $2$ \\
			
			width $k$  & $-$ &  $-$  & $2$ &  $4$  & $4$ & $4$  & $4$ &  $2$ & $4$  & $4$ &  $2$\\
			
			Conv Operator  & GIN &  GIN  & PNAConv &  PNAConv  & NNConv & GIN &  GIN  & GIN  & GIN &  GIN  & GIN \\
			
	\end{tabular}}
\end{table*}

\subsection{Datasets}

\noindent\textbf{TRIANGLES:}

The dataset TRIANGLES~\citep{knyazev2019understanding} consists of $45000$ graphs with the task of counting the number of triangles in the graph. The number of classes is $10$. We use node degrees as node features. The data splits are as follows, training ($30000$), validation ($5000$), test-original ($5000$) and test-large ($5000$). Except test-large, all data splits have smaller graphs with $N < 25$ nodes. The test-large set has $25 < N < 100$ nodes and tests the generalization ability of the model. Figure~\ref{fig:triangles} shows example graphs from the two test sets where both have same number of triangles, $Y = 6$. It can be seen in Table~\ref{tab:Trianles} that there is significant improvement in the Test-large set with increasing IR layers and width of each IR-layer suggesting better generalization to larger graphs by our model.

\noindent\textbf{Circulant Skip Links (CSL) :}
\begin{figure}[t]
	\centering
	\begin{minipage}{0.7\textwidth}
		\begin{subfigure}[t]{0.5\textwidth}
			\includegraphics[width=0.85\textwidth]{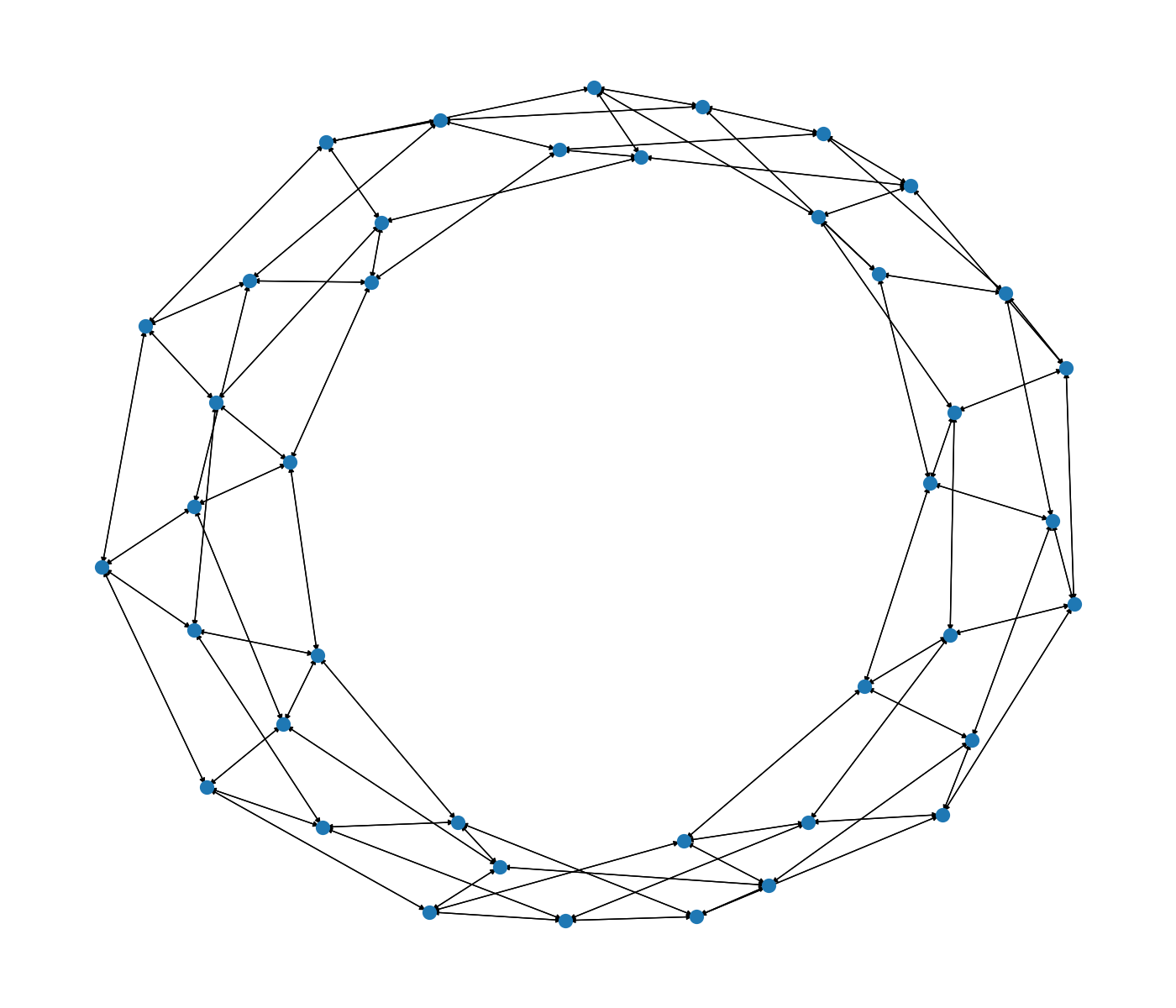}
			\label{fig:csl_a}
			\caption{$\mathcal{G}_{skip} (41, 4)$}	
		\end{subfigure}\hfil
		\begin{subfigure}[t]{0.5\textwidth}
			\includegraphics[width=0.85\textwidth]{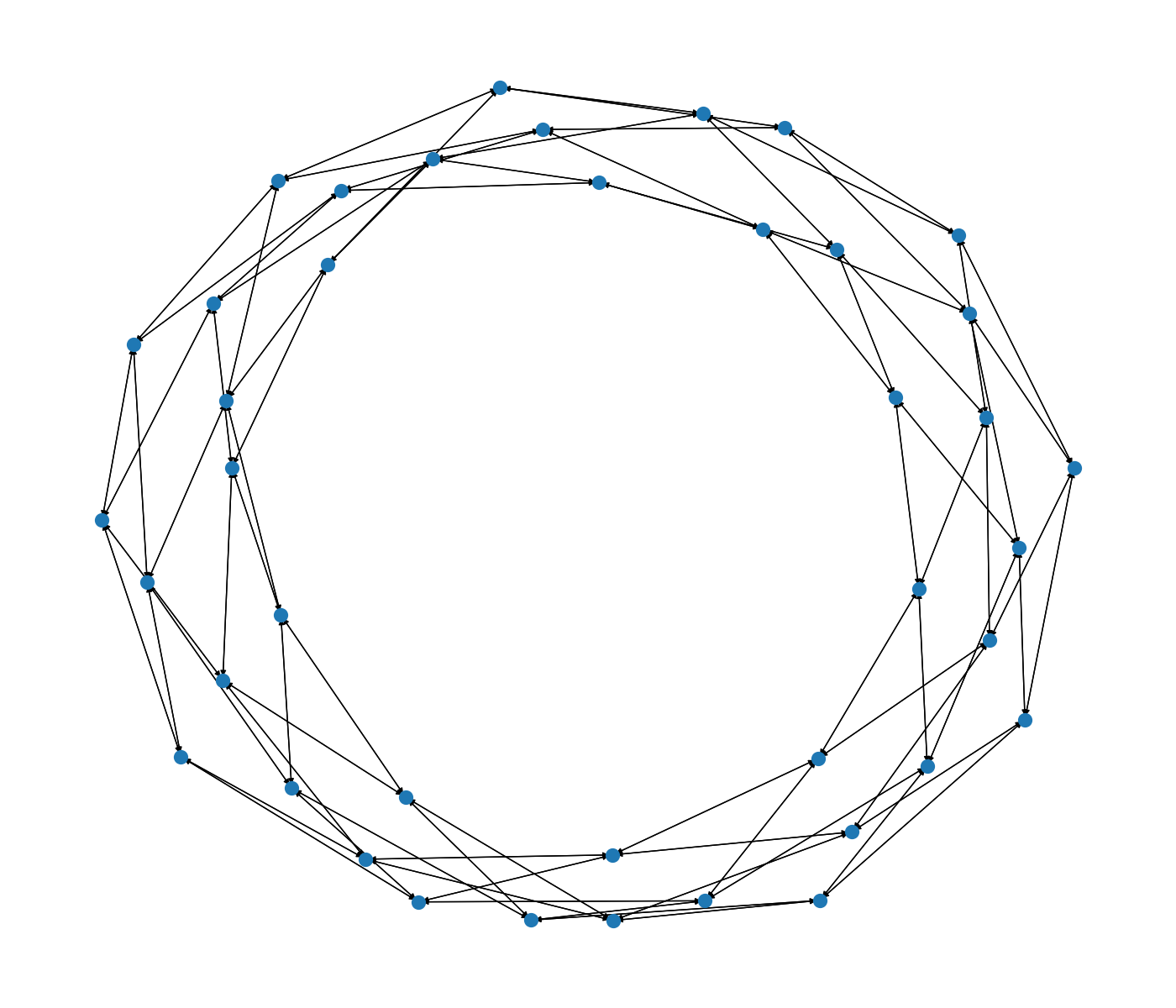}
			\label{fig:csl_b}
			\caption{$\mathcal{G}_{skip} (41, 11)$}
		\end{subfigure}
		\caption{\footnotesize Non-isomorphic $4$-regular graphs from \textsc{CSL} dataset.}
		\label{fig:csl}
	\end{minipage}\hfil
\end{figure}

The Circulant Skip Link dataset is a graph classification dataset introduced in~\cite{murphy2019relational} in order to test the expressive power of GNNs.  A CSL graph $\mathcal{G}_{skip} (M, R)$ is a 4-regular graph with $\{0,1,\dots M-1\}$ vertices and  edge between pairs of vertices which are $R$ distance away in a cyclical form. The graphs are from $10$ classes representing $R \in \{2, 3, 4, 5, 6, 9, 11, 12, 13, 16\}$. Figure~\ref{fig:csl} shows two non-isomorphic graphs from the CSL dataset with 11 nodes $R=4, 11$. The dataset has $150$ graphs with each class having $15$ graphs. Following~\citet{murphy2019relational}, we use a 5-fold cross validation split, with each split having train, validation and test data in the ratio of $3 : 1 : 1$. We use the validation split to decay the learning rate and for model selection.


\noindent\textbf{ZINC10K:}

ZINC10K dataset~\citep{jin2018junction,dwivedi2020benchmarking} contains $12000$ Zinc molecules out of which $10000$ are for training and $1000$ each for validation and test sets. The dataset comes with constrained solubility values associated with the molecules and the task is to regress over these values. The performance measure is the mean absolute error (MAE) for regression on each graph. We use the baseline results as in~\citet{morris2020weisfeiler} and add other recent better performing model PNA~\citep{corso2020principal} for a more thorough comparison. We use L1 loss and report absolute error on the test set.

\noindent\textbf{ALCHEMY10K:}

ALCHEMY10K~\citep{chen2019alchemy} is a recently released graph regression dataset. The task to regress over $12$ values related to quantum chemistry. The targets are same as in QM9 but the molecules come with more heavy atoms. We use the same smaller version of the dataset as used in~\citet{morris2020weisfeiler} with $10000$ training, $1000$ validation and $1000$ test molecules which are randomly sampled from the original full dataset. Like in ZINC10K, we add PNA~\citep{corso2020principal} for a more recent comparison. For this, we ran PNA on the same dataset with code provided officially by Pytorch-Geometric library. We use L1 loss and report MAE on the regression targets.\\

\noindent\textbf{QM9:}

QM9~\citep{ruddigkeit2012enumeration,ramakrishnan2014quantum} is a prominent large scale dataset in the domain of quantum chemistry. The dataset contains more than 130K drug-like molecules with sizes ranging from 4-29 atoms per molecule. Each molecule may contain up to 9 heavy (non-Hydrogen) atoms. The task is to regress on 12 quantum-mechanical properties associated with each molecule. In our experiments, we follow~\citet{gilmer2017neural} MPNN architecture in node and edge features. We further add edge variables for message passing and readout from both node and edge variables. For edge message passing, we simply concatenate the two node hidden vectors and pass it through an MLP and update the edge variables before finally reading out from both node and edge variables. We follow the same random split of $8:1:1$ for training, validation and testing as in ~\citet{morris2019weisfeiler,morris2020weisfeiler}. We compare with the recent GNN models as in~\citet{morris2020weisfeiler} which covers prominent GNN models tested on this dataset. Note that as in~\citet{morris2020weisfeiler}, we are not comparing with Schnet~\citep{schutt2018schnet}, Physnet~\citep{unke2019physnet} and Dimenet~\citep{klicpera2020directional} which incorporate physical knowledge in the modeling of message passing. All compared models are general GNN models. We train with L1 loss and report MAE on the targets.

\begin{figure}[t]
	\centering
	\includegraphics[width=0.55\textwidth]{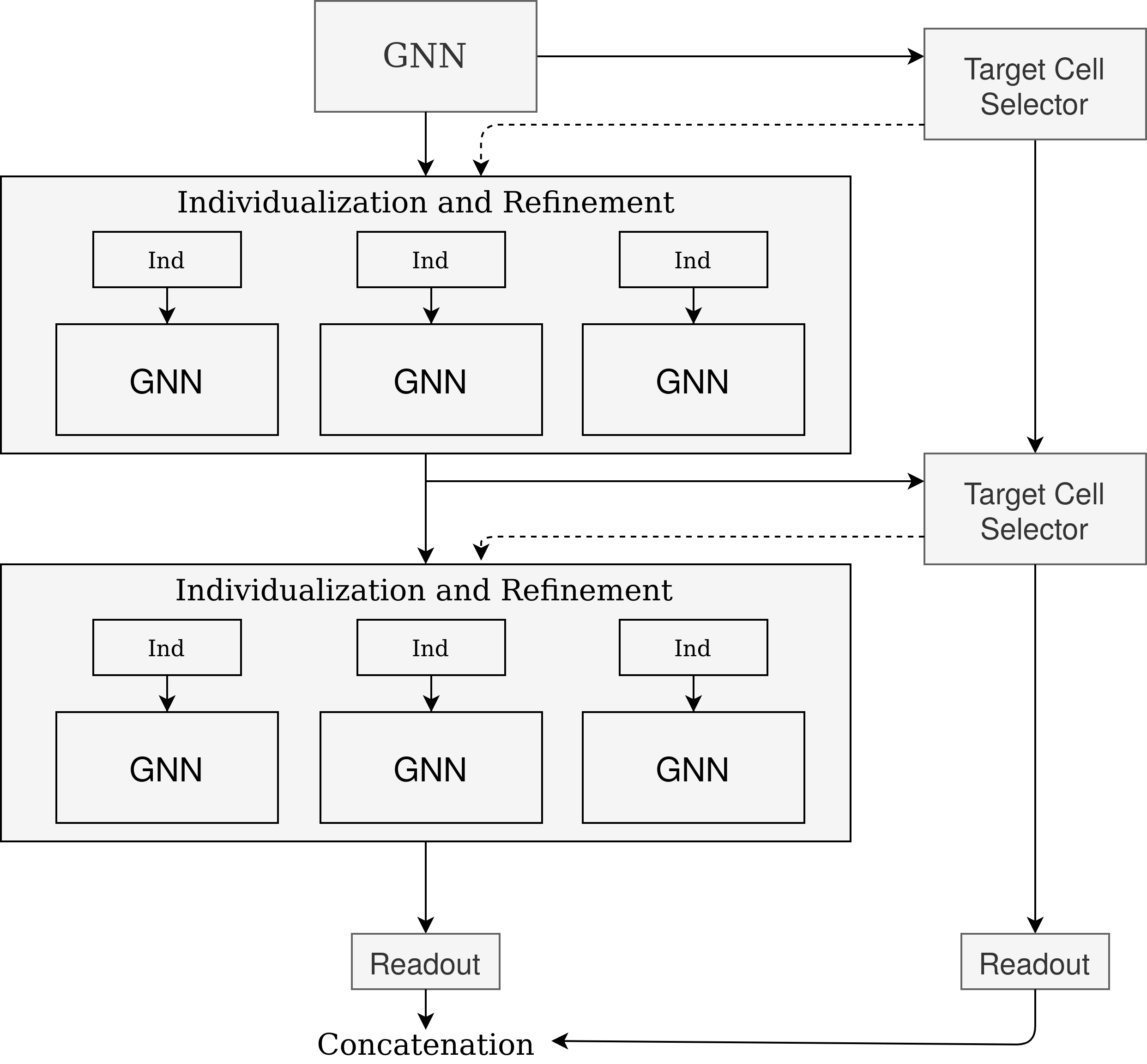}
	\caption{\footnotesize Model architecture of GNN-IR }
	\label{fig:individualization3}
\end{figure}

\end{document}